\documentclass{article}
\usepackage[group-separator={,}]{siunitx}
\usepackage{booktabs} 
\usepackage{color}
\usepackage{listings}
\usepackage{enumitem}
\usepackage{threeparttable}
\usepackage{multirow}
\usepackage{hhline}
\usepackage{subfigure}
\usepackage{array}
\usepackage{pifont}
\usepackage{booktabs}
\usepackage{dsfont}
\usepackage{microtype}
\usepackage{graphicx}
\usepackage{stmaryrd}
\usepackage{amsmath}

\usepackage{amssymb}
\usepackage{amsthm}
\usepackage{url}
\usepackage{bbm}
\usepackage{xspace}
\usepackage[normalem]{ulem}
\usepackage{booktabs, siunitx}
\usepackage[svgnames,table]{xcolor}
\usepackage[tableposition=above]{caption}
\usepackage{xcolor}
\usepackage{tkz-euclide}
\usepackage[framemethod=tikz]{mdframed}
\usepackage{empheq}
\usepackage[most]{tcolorbox}
\usepackage{adjustbox}
\usepackage{wrapfig}
\usepackage{hyperref}
\usepackage[capitalize]{cleveref}
\usepackage[noend]{algpseudocode}
\usepackage{algorithm}
\usepackage{pgfplots}
\usepgfplotslibrary{groupplots}

\usepackage{balance}
\expandafter\def\expandafter\UrlBreaks\expandafter{\UrlBreaks
  \do\a\do\b\do\c\do\d\do\e\do\f\do\g\do\h\do\i\do\j%
  \do\k\do\l\do\m\do\n\do\o\do\p\do\q\do\r\do\s\do\t%
  \do\u\do\v\do\w\do\x\do\y\do\z\do\A\do\B\do\C\do\D%
  \do\E\do\F\do\G\do\H\do\I\do\J\do\K\do\L\do\M\do\N%
  \do\O\do\P\do\Q\do\R\do\S\do\T\do\U\do\V\do\W\do\X%
  \do\Y\do\Z}

\expandafter\def\expandafter\UrlBreaks\expandafter{\UrlBreaks
  \do\a\do\b\do\c\do\d\do\e\do\f\do\g\do\h\do\i\do\j%
  \do\k\do\l\do\m\do\n\do\o\do\p\do\q\do\r\do\s\do\t%
  \do\u\do\v\do\w\do\x\do\y\do\z\do\A\do\B\do\C\do\D%
  \do\E\do\F\do\G\do\H\do\I\do\J\do\K\do\L\do\M\do\N%
  \do\O\do\P\do\Q\do\R\do\S\do\T\do\U\do\V\do\W\do\X%
  \do\Y\do\Z}

\definecolor{codegreen}{rgb}{0,0.6,0}
\definecolor{codegray}{rgb}{0.5,0.5,0.5}
\definecolor{codepurple}{rgb}{0.58,0,0.82}
\definecolor{backcolour}{rgb}{1,1,1}
 
\lstdefinestyle{mystyle}{
    backgroundcolor=\color{backcolour},   
    commentstyle=\color{codegreen},
    keywordstyle=\color{magenta},
    numberstyle=\tiny\color{codegray},
    stringstyle=\color{codepurple},
    basicstyle=\ttfamily\footnotesize,
    breakatwhitespace=false,         
    breaklines=true,                 
    captionpos=b,                    
    keepspaces=true,                 
    numbers=left,                    
    numbersep=5pt,                  
    showspaces=false,                
    showstringspaces=false,
    showtabs=false,           
    tabsize=2,
    xleftmargin=1.5em
}

\newtheorem{lemma}{Lemma}[section]
\newtheorem{theorem}{Theorem}[section]

\newtheorem{example}{Example}[section]
\newtheorem{remark}{Remark}[section]

\newcommand{\technique}{BagFlip\xspace}
\newcommand{\perturb}{\pi}

\newcommand{\bfx}{\mathbf{x}}

\newcommand{\poisoned}[1]{\widetilde{#1}}
\newcommand{\poisonfeat}{s}
\newcommand{\poisonins}{r}

\newcommand{\outcome}{o}
\newcommand{\subspace}{\mathcal{L}}
\newcommand{\secondy}{y'}
\newcommand{\firsty}{y^*}
\newcommand{\secondp}{p'}
\newcommand{\firstp}{p^*}
\newcommand{\dis}[2]{\|#1-#2\|_0}
\newcommand{\featflip}{\textsc{f}_{\poisonfeat}}
\newcommand{\featflipone}{\textsc{f}_{1}}
\newcommand{\featflipinf}{\textsc{f}_{\infty}}
\newcommand{\labelflip}{\textsc{l}}

\newcommand{\featfliplabelflip}{\textsc{fl}_{\poisonfeat}}
\newcommand{\featfliplabelflipone}{\textsc{fl}_{1}}
\newcommand{\featfliptwo}{\textsc{f}_{2}}
\newcommand{\featflipfour}{\textsc{f}_{4}}
\newcommand{\featflipeight}{\textsc{f}_{8}}

\newcommand{\selectflip}{\mu}

\newcommand{\bfxtest}{\bfx}
\newcommand{\ytest}{y}
\newcommand{\alg}{A}
\newcommand{\ualg}{A^?}
\newcommand{\salg}{\bar{\alg}}
\newcommand{\sualg}{\bar{\ualg}}
\newcommand{\modelpredo}[1]{\alg(#1)}
\newcommand{\umodelpredo}[1]{\ualg(#1)}
\newcommand{\modelpred}[2]{\alg(#1,#2)}
\newcommand{\smodelpred}[2]{\salg(#1,#2)}
\newcommand{\poisonset}[1]{S^{#1}_r(D)}
\newcommand{\lb}{\mathrm{lb}}
\newcommand{\ub}{\mathrm{ub}}
\newcommand{\sample}[1]{\dot{#1}}
\newcommand{\poisondist}{\poisoned{\mu}}
\newcommand{\cleandist}{\mu}

\newcommand{\assumeradii}{R}
\newcommand{\symbolremain}{\rho}
\newcommand{\symbolflip}{\gamma}
\newcommand{\indexset}{B}
\newcommand{\smallk}{\kappa}
\newcommand{\diffu}{\Delta}
\newcommand{\diffv}{\poisoned{\Delta}}
\newcommand{\cleandistmass}[1]{p_{\cleandist}(#1)}
\newcommand{\poisondistmass}[1]{p_{\poisondist}(#1)}

\newcommand{\PP}{\mathrm{Pr}}
\newcommand{\KL}{\mathrm{KL}}

\PassOptionsToPackage{numbers, compress}{natbib}

\usepackage[final]{neurips_2022}

\usepackage[utf8]{inputenc} 
\usepackage[T1]{fontenc}    
\usepackage{hyperref}       
\usepackage{url}            
\usepackage{booktabs}       
\usepackage{amsfonts}       
\usepackage{nicefrac}       
\usepackage{microtype}      
\usepackage{xcolor}         

\title{BagFlip: A Certified Defense against Data Poisoning}

\author{%
   Yuhao Zhang \\
   University of Wisconsin-Madison \\
   \texttt{yuhaoz@cs.wisc.edu}
   \And
   Aws Albarghouthi \\
   University of Wisconsin-Madison \\
   \texttt{aws@cs.wisc.edu} \\
   \AND
   Loris D'Antoni \\
   University of Wisconsin-Madison \\
   \texttt{loris@cs.wisc.edu} \\
}

\begin{document}

\maketitle

\begin{abstract}
Machine learning models are vulnerable to data-poisoning attacks, in which an attacker maliciously modifies the training set to change the prediction of a learned model. 
In a \emph{trigger-less} attack, the attacker can modify the training set but not the test inputs, while in a \emph{backdoor} attack the attacker can also modify test inputs. 
Existing model-agnostic defense approaches either cannot handle backdoor attacks or do not provide effective certificates (i.e., a proof of a defense).
We present \technique, a model-agnostic certified approach that can effectively defend against both trigger-less and backdoor attacks. 
We evaluate \technique on image classification and malware detection datasets. 
\technique is equal to or more effective than the state-of-the-art approaches for trigger-less attacks and more effective than the state-of-the-art approaches for backdoor attacks. 
\end{abstract}

\section{Introduction}
Recent works have shown that machine learning models are vulnerable to data-poisoning attacks, where the attackers maliciously modify the training set to influence the prediction of the victim model as they desire.
In a \emph{trigger-less} attack~\cite{triggerless_convpolytope, triggerless_featurecollision, triggerless_SpamFilter, triggerless_SVM, triggerless_bullseyepolytope, triggerless_witchbrew}, the attacker can modify the training set but not the test inputs, while in a \emph{backdoor} attack~\cite{backdoor_hidden, backdoor_labelconsistent, backdoor_latent, backdoor_Malwarepoison, backdoor_targetDL, backdoor_trojan, backdoor_NLP, backdoor_code} the attacker can also modify test inputs.
Effective attack approaches have been proposed for various domains such as image recognition~\cite{badnets}, sentiment analysis~\cite{backdoor_NLP}, and malware detection~\cite{backdoor_Malwarepoison}. 

Consider the malware detection setting.
An attacker can modify the training data by adding a special signature---e.g., a line of code---to a set of benign programs.
The idea is to make the learned model correlate the presence of the signature with benign programs.
Then, the attacker can sneak a piece of malware past the model by including the signature,
fooling the model into thinking it is a benign program.
Indeed, it has been shown that if the model is trained on a dataset with only a few poisoned examples, a backdoor signature can be installed in the learned model~\cite{backdoor_Malwarepoison, backdoor_code}.
Thus, data-poisoning attacks are of great concern to the safety and security of machine learning models and systems, particularly as training data is gathered from different sources, e.g., via web scraping.

Ideally, a defense against data poisoning should fulfill the following desiderata: (1) Construct \textbf{effective certificates} (proofs) of the defense. (2) Defend against \textbf{both trigger-less and backdoor attacks}. (3) Be \textbf{model-agnostic}.
It is quite challenging to fulfill all three desiderata; indeed, existing techniques are forced to make tradeoffs.
For instance, empirical approaches~\cite{heu_advtrain, heu_featureselect, heu_finetune, heu_gendist, heu_neuralcleanse, heu_robustcovariance, heu_spectral, heu_strip} cannot construct certificates and are likely to be bypassed by new attack approaches~\cite{break_attackdefense, break_backdoorfl, break_sanitization}.
Some certified approaches~\cite{RAB, featureflip}  provide \textit{ineffective} certificates for both trigger-less and backdoor attacks.
Some certified approaches~\cite{bagging, partitioning, framework_def, labelflip, differential, dpa_improve} provide effective certificates for trigger-less attacks but not backdoor attacks. %
Other certification approaches~\cite{certkrNN, decisiontree, prog_decisiontree} are restricted to specific learning algorithms, e.g., decision trees.

This paper proposes \technique, a \textit{model-agnostic certified} approach that uses \emph{randomized smoothing}~\cite{randomizesmoothing, divergences, tightflipbound} to effectively defend against both trigger-less and backdoor attacks~(\Cref{sec:smoothing}). 
\technique uses a novel smoothing distribution that combines \emph{bagging} of the training set and noising of the training data and the test input by randomly \emph{flipping} features and labels.

Although both bagging-based and noise-based approaches have been proposed independently in the literature, combining them makes it challenging to compute the \emph{certified radius}, i.e., the amount of poisoning a learning algorithm can withstand without changing its prediction.

To compute the certified radius precisely, we apply the Neyman--Pearson lemma to the sample space of \technique's smoothing distribution. 
This lemma requires partitioning the outcomes in the sample space into subspaces such that the likelihood ratio in each subspace is a constant.
However, because our sample space is exponential in the number of features, we cannot naively apply the lemma.
To address this problem, we exploit properties of our smoothing distribution to design an efficient algorithm that partitions the sample space into polynomially many subspaces (\Cref{sec:precise_approach}).
Furthermore, we present a \emph{relaxation} of the Neyman–Pearson lemma that further speeds up the computation (\Cref{sec:imprecise}).
Our evaluation against existing approaches shows that \technique is comparable to them or more effective for trigger-less attacks and more effective for backdoor attacks (\Cref{sec: experiment}). 

\section{Related Work}
\label{sec: related_work}

Our \textit{model-agnostic} approach \technique uses randomized smoothing to compute \textit{effective certificates} for \textit{both} trigger-less and backdoor attacks.
\technique focuses on feature-and-label-flipping perturbation functions that modify training examples and test inputs.
We discuss how existing approaches differ.

Some model-agnostic certified approaches can defend against both trigger-less and backdoor attacks but cannot construct effective certificates.
\citet{featureflip, RAB} defended feature-flipping poisoning attacks and $l_2$-norm poisoning attacks, respectively.
However, their certificates are practically ineffective due to the curse of dimensionality~\cite{curse}.

Some model-agnostic certified approaches construct effective certificates but cannot defend against backdoor attacks.
\citet{bagging} proposed to defend against \textit{general} trigger-less attacks, i.e., the attackers can add/delete/modify examples in the training dataset, by bootstrap aggregating (bagging).
\citet{framework_def} extended bagging by designing other selection strategies, e.g., selecting without replacement and with a fixed probability.
\citet{labelflip} defended against label-flipping attacks and instantiated their framework on linear classifiers.
Differential privacy~\cite{differential} can also provide probabilistic certificates for trigger-less attacks, but it cannot handle backdoor attacks, and its certificates are ineffective.
\citet{partitioning} proposed a deterministic partition aggregation~(DPA) to defend against general trigger-less attacks by partitioning the training dataset using a secret hash function. 
\citet{dpa_improve} further improved DPA by introducing a spread stage.

Other certified approaches construct effective certificates but are not model-agnostic.
\citet{certkrNN} provided deterministic certificates \textit{only} for nearest neighborhood classifiers~(kNN/rNN), but for both trigger-less and backdoor attacks.
\citet{prog_decisiontree, decisiontree} provided deterministic certificates \textit{only} for decision trees but against general trigger-less attacks.


\section{Problem Definition}
\label{sec: prob_def}

We take a holistic view of training and inference as a single deterministic algorithm $\alg$.
Given a dataset $D = \{(\bfx_1, y_1), \ldots, (\bfx_n, y_n)\}$
%
and a (test) input $\bfxtest$, we write $\modelpred{D}{\bfxtest}$ to denote the prediction of algorithm $\alg$ on the input $\bfxtest$ after being trained on dataset $D$.

We are interested in certifying that the algorithm will still behave ``well'' after training on a tampered dataset.
Before describing what ``well'' means, to define our problem we need to assume a \emph{perturbation space}---i.e., what possible changes the attacker could make to the dataset.
Given a pair $(\bfx, y)$, we write $\perturb(\bfx, y)$ to denote the set of perturbed examples that an attacker can transform the example $(\bfx, y)$ into. 
Given a dataset $D$ and a \textit{radius} $\poisonins\geq 0$, we define the \textit{perturbation space} as the set of datasets that can be obtained by modifying up to $\poisonins$ examples in $D$ using the perturbation $\perturb$:
\[
\poisonset{\perturb} = \left\{\{(\poisoned{\bfx}_i, \poisoned{y}_i)\}_i ~\bigg|~ \forall i.\ (\poisoned{\bfx}_i, \poisoned{y}_i) \in \perturb(\bfx_i, y_i),\ \sum_{i=1}^n \mathds{1}_{(\poisoned{\bfx}_i, \poisoned{y}_i) \neq (\bfx_i,y_i)} \le r\right\}
\]

\textbf{Threat models.}
We consider two attacks: one where an attacker can perturb only the training set (a \emph{trigger-less} attack) and one where the attacker can perturb both the training set and the test input (a \emph{backdoor} attack). We assume a backdoor attack scenario where test input perturbation is the same as training one.
If these two perturbation spaces can perturb examples differently, we can always over-approximate them by their union.
In the following definitions, we assume that we are given a perturbation space $\perturb$, a radius $\poisonins\geq 0$, and a benign training dataset $D$ and test input $\bfx$.

%

We say that algorithm $\alg$ is robust to a \textbf{trigger-less attack} on the test input $\bfxtest$ if $\alg$  yields the same prediction on the input $\bfxtest$ when trained on any perturbed dataset $\poisoned{D}$ and the benign dataset $D$. Formally,
\begin{align}
    \forall \poisoned{D} \in \poisonset{\perturb}.\ \modelpred{\poisoned{D}}{\bfxtest} = \modelpred{D}{\bfxtest}\label{eq:trigger_less_goal}
\end{align}

We say that algorithm $\alg$ is robust to a \textbf{backdoor attack} on the test input $\bfxtest$ 
if the algorithm trained on any perturbed dataset $\poisoned{D}$ produces the prediction $\modelpred{D}{\bfxtest}$ 
on any perturbed input $\poisoned{\bfxtest}$. 
Let $\perturb(\bfxtest, \ytest)_1$ denote the projection of the perturbation space $\perturb(\bfxtest, \ytest)$ onto the feature space (the attack can only backdoor features of the test input).
Robustness to a backdoor attack is defined as
\begin{align}
     \forall \poisoned{D} \in \poisonset{\perturb},\ \poisoned{\bfxtest} \in \perturb(\bfxtest, \ytest)_1.\ \modelpred{\poisoned{D}}{\poisoned{\bfxtest}} = \modelpred{D}{\bfxtest}\label{eq:goal}
\end{align}

Given a large enough radius $r$, an attacker can always change enough examples and succeed at breaking robustness for either kind of attack. 
Therefore, we will focus on computing the maximal radius $r$, for which we can prove that Eq~\ref{eq:trigger_less_goal}~and~\ref{eq:goal} hold for a given  perturbation function $\perturb$.
We refer to this quantity as the \textit{certified radius}.
It is infeasible to prove Eq~\ref{eq:trigger_less_goal}~and~\ref{eq:goal} by enumerating all possible $\poisoned{D}$ because $\poisonset{\perturb}$ can be ridiculously large, e.g., $|\poisonset{\perturb}| > 10^{30}$ when $|D|=1000$ and $\poisonins=10$.

\textbf{Defining the perturbation function.}
We have not yet specified the perturbation function $\perturb$, i.e., how the attacker can modify examples.
In this paper, we focus on the following perturbation spaces.

Given a bound $\poisonfeat \geq 0$, a \textbf{feature-and-label-flipping perturbation}, $\featfliplabelflip$, allows the attacker to modify the values of up to $\poisonfeat$ features and the label in an example $(\bfx, y)$, where $\bfx \in [K]^d$ (i.e., $\bfx$ is a $d$-dimensional feature vector with each dimension having $\{0,1,\ldots, K\}$ categories). Formally,
    \begin{align*}
            \featfliplabelflip(\bfx, y) = \{(\poisoned{\bfx}, y')\mid \|\bfx - \poisoned{\bfx}\|_0 + \mathds{1}_{y\neq y'} \le \poisonfeat\}, 
    \end{align*}
There are two special cases of $\featfliplabelflip$, a \textbf{feature-flipping perturbation} $\featflip$ and a \textbf{label-flipping perturbation} $\labelflip$.
Given a bound $\poisonfeat \geq 0$, $\featflip$ allows the attacker to modify the values of up to $\poisonfeat$ features in the input $\bfx$ but not the label.
$\labelflip$ only allows the attacker to modify the label of a training example. Note that $\labelflip$ cannot modify the test input's features, so it can only be used in trigger-less attacks.

\begin{example}
\label{exp: perturbation}
If $D$ is a binary-classification image dataset, where each pixel is either black or white, then the perturbation function $\featflipone$ assumes the attacker can modify up to one pixel per image. 
\end{example}

    
The goal of this paper is to design a certifiable algorithm that can defend against trigger-less and backdoor attacks (Section~\ref{sec:smoothing}) by computing the certified radius (Sections~\ref{sec:precise_approach}~and~\ref{sec:imprecise}).
Given a benign dataset $D$, our algorithm certifies that an attacker can perturb $D$ by some amount (the certified radius) 
without changing the prediction.
Symmetrically, if we suspect that $D$ is poisoned, our algorithm certifies that even if an attacker had not poisoned $D$ by up to the certified radius, the prediction would have been the same.
The two views are equivalent, and we use the former in the paper.
\section{\technique: Dual-Measure Randomized Smoothing}
\label{sec:smoothing}

Our approach, which we call \technique, is a model-agnostic certification technique.
Given a learning algorithm $\alg$, we want to automatically construct a new learning algorithm $\salg$ with certified poisoning-robustness guarantees (\cref{eq:trigger_less_goal,eq:goal}).
To do so, we adopt and extend the framework of \emph{randomized smoothing}.
Initially used for test-time robustness, randomized smoothing robustifies a function $f$ by carefully constructing a \emph{noisy} version $\bar{f}$ and theoretically analyzing the guarantees of $\bar{f}$.

Our approach, \technique, constructs a noisy algorithm $\salg$ by randomly
perturbing the training set and the test input and invoking $\alg$ on the result.
Formally, if we have a set of output labels $\mathcal{C}$, we define the \emph{smoothed} learning algorithm $\salg$ as follows:
\begin{align}
\smodelpred{D}{\bfxtest} \triangleq \underset{y\in \mathcal{C}}{\mathrm{argmax}} \underset{\sample{D}, \sample{\bfxtest} \sim \mu(D, \bfxtest)}{\PP}(\modelpred{\sample{D}}{\sample{\bfxtest}}=y), \label{eq:smoothing1}    
\end{align}
where $\mu$ is a carefully designed probability distribution---the \emph{smoothing distribution}---%
over the training set and the test input ($\sample{D}, \sample{\bfxtest}$ are a sampled dataset and a test input from the smoothing distribution).
 

We present two key contributions that enable \technique to efficiently certify robustness up to large poisoning radii.
First, we define a smoothing distribution $\selectflip$ that combines \emph{bagging}~\cite{bagging} of the training set $D$, and noising~\cite{featureflip} of the training set and the test input $\bfx$ (done by randomly \emph{flipping} features and labels). This combination, described below, allows us to defend against trigger-less and backdoor attacks.
However, this combination also makes it challenging to compute the certified radius
due to the combinatorial explosion of the sample space.
Our second contribution is a partition strategy for the Neyman--Pearson lemma that results in an efficient certification algorithm (\cref{sec:precise_approach}), as well as a relaxation of the Neyman--Pearson lemma that further speeds up certification (\cref{sec:imprecise}).

\textbf{Smoothing distribution.}
We describe the smoothing distribution $\selectflip$ that defends against feature and label flipping.
Given a dataset $D = \{(\bfx_i, y_i)\}$ and a test input $\bfxtest$, sampling from 
$\selectflip(D,\bfx)$ generates a random dataset and test input in the following way:
(1) Uniformly select $k$ examples from $D$ with replacement and record their indices as $w_1, \ldots, w_k$. (2) 
Modify each selected example $(\bfx_{w_i},y_{w_i})$ and the test input $\bfxtest$ to $(\bfx_{w_i}',y_{w_i}')$ and $\bfxtest'$, respectively, as follows: For each feature, with probability $1-\symbolremain$,
uniformly change its value to one of the other categories in $\{0,\ldots,K\}$.
Randomly modify labels $y_{w_i}$ in the same way.
In other words, each feature will be \emph{flipped} to another value from the domain 
with probability $\symbolflip \triangleq \frac{1 - \symbolremain}{K}$,
where $\symbolremain \in [0, 1]$ is the parameter controlling the noise level. 

For $\featflip$ (resp. $\labelflip$), we simply do not modify the labels (resp. features) of
the $k$ selected examples. 
\begin{example}
\label{exp: distribution}
Suppose we defend against $\featflipone$ in a trigger-less setting, then the distribution $\selectflip$ will not modify labels or the test input. Let $\symbolremain = \frac{4}{5}$, $k=1$.
Following \Cref{exp: perturbation}, let the binary-classification image dataset $D=\{(\bfx_1, y_1), (\bfx_2, y_2)\}$, 
where each image contains only one pixel. 
Then, one possible element of $\selectflip(D, \bfxtest)$ can be the pair $(\{(\bfx_2', y_2)\}, \bfxtest)$, where $\bfx_2' = \bfx_2$.
The probability of this element is $\frac{1}{2} \times \symbolremain = 0.4$ because we uniformly select one out of two examples and do not flip any feature, that is, the single feature retains its original value. 
\end{example}

\section{A Precise Approach to Computing the Certified Radius}
\label{sec:precise_approach}

In this section, we show how to compute the certified radius of the smoothed algorithm $\salg$ given a dataset $D$, a test input $\bfxtest$, and a perturbation function $\perturb$.
We focus on binary classification and provide the multi-class case in \cref{sec: lemma_multi_class}.

Suppose that we have computed the prediction $\firsty=\smodelpred{D}{\bfxtest}$.
We want to show how many examples we can perturb in $D$ to obtain any other $\poisoned{D}$ so the prediction remains $\firsty$.
Specifically, we want to find the largest possible radius $\poisonins$  
such that 
\begin{align}
    \forall \poisoned{D} \in \poisonset{\perturb},\ \poisoned{\bfxtest} \in \perturb(\bfxtest, \ytest)_1.\ \underset{\sample{D}, \sample{\bfxtest} \sim \mu(\poisoned{D},\poisoned{\bfxtest})}{\PP}(\modelpred{\sample{D}}{\sample{\bfxtest}}=\firsty) > 0.5 \label{eq:certgoal1}
\end{align}
We first show how to \textit{certify} that Eq~\ref{eq:certgoal1} holds for a given $\poisonins$ and then rely on binary search to compute the largest $\poisonins$, i.e., the certified radius.\footnote{It is difficult to get a closed-form solution of the certified radius (as done in~\cite{bagging, RAB}) in our setting because the distribution of \technique is complicated. We rely on binary search as it is also done in \citet{featureflip, tightflipbound}. \Cref{sec: kl_bound} shows a \textit{loose}  closed-form bound on the certified radius by KL-divergence~\cite{divergences}.}
In \Cref{sec: lemma}, we present the Neyman--Pearson lemma to certify Eq~\ref{eq:certgoal1} as it is a common practice in randomized smoothing. 
In \Cref{sec: featflip}, we show how to compute the certified radius for the distribution $\selectflip$ and the perturbation functions $\featflip, \featfliplabelflip$, and $\labelflip$.

\subsection{The Neyman--Pearson Lemma}
\label{sec: lemma}

Hereinafter, we simplify the notation and use $\outcome$ to denote the pair $(\sample{D},\sample{\bfxtest})$ and $\modelpredo{\outcome}$ to denote the prediction of the algorithm training and evaluating on $\outcome$.
We further simplify the distribution $\mu(D,\bfxtest)$ as $\cleandist$ and the distribution $\mu(\poisoned{D},\poisoned{\bfxtest})$ as $\poisondist$.
We define the performance of the smoothed algorithm $\salg$ on dataset $D$, i.e., the probability of predicting $\firsty$, as $\firstp = \PP_{\outcome \sim \cleandist}(\modelpredo{\outcome}=\firsty)$.

The challenge of certifying Eq~\ref{eq:certgoal1} is that we cannot directly estimate the performance of the smoothed algorithm on the perturbed data, i.e., $\PP_{\outcome \sim \poisondist}(\modelpredo{\outcome}=\firsty)$, because $\poisondist$ is universally quantified. 
To address this problem, we use the Neyman--Pearson lemma to find a lower bound $\lb$ for $\PP_{\outcome \sim \poisondist}(\modelpredo{\outcome}=\firsty)$.
We do so by constructing a worst-case algorithm $\sualg$ and distribution $\poisondist$.
Note that we add the superscript $?$ to denote a worst-case algorithm.
Specifically, we minimize $\PP_{\outcome \sim \poisondist}(\umodelpredo{\outcome}=\firsty)$ while maintaining the algorithm's performance on $\cleandist$, i.e., keeping $\PP_{\outcome \sim \cleandist}(\umodelpredo{\outcome}=\firsty) = \firstp$.
We use $\mathcal{A}$ to denote the set of all possible algorithms.
We formalize the computation of the lower bound $\lb$ as the following constrained minimization objective:  
\begin{small}
\begin{align}
    \lb \triangleq \min_{\sualg\in \mathcal{A}} \underset{\outcome \sim \poisondist}{\PP}(\umodelpredo{\outcome}=\firsty)\label{eq:lbub}
    \ \ \ \ s.t.&\ \underset{\outcome \sim \cleandist}{\PP}(\umodelpredo{\outcome}=\firsty)=\firstp
    \text{ and }  \poisoned{D} \in \poisonset{\perturb}, \poisoned{\bfxtest} \in \perturb(\bfxtest, \ytest)_1 
\end{align}
\end{small}%
It is easy to see that $\lb$ is the lower bound of $\PP_{\outcome \sim \poisondist}(\modelpredo{\outcome}=\firsty)$ in Eq.~\ref{eq:certgoal1} because $\salg \in \mathcal{A}$ and $\salg$ satisfies the minimization constraint.
Thus, $\lb > 0.5$ implies the correctness of Eq.~\ref{eq:certgoal1}.

We show how to construct $\sualg$ and $\poisondist$ greedily.
For each outcome $\outcome=(\sample{D},\sample{\bfxtest})$ in the sample space $\Omega$---i.e., the set of all possible sampled datasets  and test inputs---we define the likelihood ratio of $\outcome$ as $\eta(\outcome) = p_{\cleandist}(\outcome) / p_{\poisondist}(\outcome)$, where $p_{\cleandist}$ and $p_{\poisondist}$ are the PMFs of $\cleandist$ and $\poisondist$, respectively.

The key idea is as follows: We partition $\Omega$ into finitely many disjoint subspaces $\subspace_1, \ldots, \subspace_m$ such that the likelihood ratio in each subspace $\subspace_i$ is some constant $\eta_i \in [0,\infty]$, i.e., $\forall \outcome \in \subspace_i. \eta(\outcome) = \eta_i$.
We can sort and reorder the subspaces by likelihood ratios such that $\eta_1 \ge \ldots \ge \eta_m$. 
We denote the probability mass of $\cleandist$ on subspace $\subspace_i$ as $\cleandistmass{\subspace_i}$.

\begin{example}
\label{exp: merge}
Suppose $\cleandistmass{\outcome_1} {=} \frac{4}{10}, \poisondistmass{\outcome_1} {=} \frac{4}{10}, \cleandistmass{\outcome_2} {=} \frac{1}{10}, \poisondistmass{\outcome_2} {=} \frac{1}{10}, \cleandistmass{\outcome_3} {=} \frac{4}{10}, \poisondistmass{\outcome_3} {=} \frac{1}{10}$, $\cleandistmass{\outcome_4} {=} \frac{1}{10}, \poisondistmass{\outcome_4} {=} \frac{4}{10}$.
We can partition $\outcome_1$ and $\outcome_2$ into one subspace $\subspace$ because $\eta(\outcome_1) {=} \eta(\outcome_2) {=} 1$. 
\end{example}

The construction of the $\sualg$ that minimizes Eq.~\ref{eq:lbub} is a greedy process, which iteratively assigns $\umodelpredo{\outcome} = \firsty$ for $\subspace_1, \subspace_2, \ldots$ until the budget $\firstp$ is met.
The worst-case $\poisondist$ can also be constructed greedily by maximizing the top-most likelihood ratios, and we can prove that the worst-case happens when the difference between $\cleandist$ and $\poisondist$ is maximized, i.e., $\poisoned{D}$ and $\poisoned{\bfx}$ are maximally perturbed.
The following theorem adapts the Neyman--Pearson lemma to our setting.

\begin{theorem}[Neyman--Pearson Lemma for $\featfliplabelflip, \featflip, \labelflip$]
\label{thm:np_bound}
Let $\poisoned{D}$ and $\poisoned{\bfx}$ be a maximally perturbed dataset and test input, i.e., $|\poisoned{D} \setminus D|=\poisonins$, $\dis{\poisoned{\bfxtest}}{\bfxtest} = \poisonfeat$, and $\dis{\poisoned{\bfxtest}_i}{\bfxtest_i} + \mathds{1}_{\poisoned{y}_i \neq y_i} = \poisonfeat$, for each perturbed example $(\poisoned{\bfxtest}_i, \poisoned{y}_i)$ in $\poisoned{D}$.
Let $i_\lb \triangleq \mathrm{argmin}_{{i \in [1,m]}} \sum_{j=1}^i \cleandistmass{\subspace_j} \ge \firstp$. 
Then, 
$\lb = \sum_{j=1}^{i_\lb - 1} \poisondistmass{\subspace_j} + \left(\firstp - \sum_{j=1}^{i_\lb - 1} \cleandistmass{\subspace_j}\right)/\eta_{i_\lb}$.
\end{theorem}

We say that $\lb$ is \emph{tight} due to the existence of the minimizer $\sualg$ and $\poisondist$ (see  \Cref{sec: lemma_multi_class}).

\begin{example}
\label{exp: lemma}
Following \Cref{exp: merge}, suppose $\Omega = \{\outcome_1, \ldots, \outcome_4\}$ and we partition it into $\subspace_1 = \{\outcome_3\}, \subspace_2 = \{\outcome_1, \outcome_2\}$, $\subspace_3 = \{\outcome_4\}$, and $\eta_1 = 4, \eta_2 = 1, \eta_3 = 0.25$.
Let $\firstp=0.95$, then we have $i_\lb = 3$ and $\lb = 0.1 + 0.5 + (0.95-0.4-0.5) / 0.25 = 0.8$.
\end{example}

\subsection{Computing the Certified Radius of \technique}
\label{sec: featflip}
Computing the certified radius boils down to computing $\lb$ in Eq~\ref{eq:lbub} using Theorem~\ref{thm:np_bound}.
To compute $\lb$ for \technique, we address the following two challenges: 
1) The argmin of computing $i_\lb$ and the summation of $\lb$ in Theorem~\ref{thm:np_bound} depend on the number of subspaces $\subspace_j$'s.
    We design a \textbf{partition strategy} that partitions $\Omega$ into a polynomial number of subspaces~($O(k^2d)$).
    Recall that $k$ is the size of the bag sampled from $D$
    and $d$ is the number of features.
2) In Theorem~\ref{thm:np_bound}, $\lb$ depends on $\cleandistmass{\subspace_j}$, which according to its definition~(Eq~\ref{eq:prd'}) can be computed in exponential time~($O(kd^{2k})$).
    We propose an \textbf{efficient algorithm} that computes these two quantities in polynomial time~($O(d^3+k^2d^2)$).
    
We first show how to address these challenges for $\featflip$ and then show how to handle $\featfliplabelflip$ and $\labelflip$.

\textbf{Partition strategy.}
We partition the large sample space $\Omega$ into disjoint subspaces $\subspace_{c,t}$ that depend on $c$, the number of perturbed examples in the sampled dataset, and $t$, the number of features the clean data and the perturbed data differ on with respect to the sampled data.
Intuitively, $c$ takes care of the bagging distribution and $t$ takes care of the feature-flipping distribution.
Formally, 
\begin{small}
\begin{align}
    \subspace_{c,t} = \{&(\{(\bfx_{w_i}', y_{w_i})\}_i, \bfx')\mid \nonumber\\
    &\sum_{i=1}^k \mathds{1}_{\bfx_{w_i} \neq \poisoned{\bfx}_{w_i}}=c ,\label{eq:Ldef1} \\
    &\underbrace{\left(\sum_{i=1}^k \dis{\bfx_{w_i}'}{\bfx_{w_i}} + \dis{\bfx'}{\bfxtest}\right)}_{\diffu} - \underbrace{\left(\sum_{i=1}^k \dis{\bfx_{w_i}'}{\poisoned{\bfx}_{w_i}} + \dis{\bfx'}{\poisoned{\bfxtest}}\right)}_{\diffv} = t\}\label{eq:Ldef3}
\end{align}
\end{small}%
Eq~\ref{eq:Ldef1} means that there are $c$ perturbed indices sampled in $\outcome$.
$\diffu$ (and $\diffv$) in Eq~\ref{eq:Ldef3} counts how many features the sampled and the clean data (the perturbed data) differ on.
The number of possible subspaces $\subspace_{c,t}$, which are disjoint by definition, is $O(k^2d)$ because $0\le c \le k$ and $|t| \le (k+1)d$.

The next theorems show how to compute the likelihood ratio of $\subspace_{c,t}$ and $\cleandistmass{\subspace_{c, t}}$.

\begin{theorem}[Compute $\eta_{c, t}$]
\label{thm:selectflip_main1}
$\eta_{c,t} = \cleandistmass{\subspace_{c, t}}/ \poisondistmass{\subspace_{c, t}}  = (\symbolflip / \symbolremain)^{t}$, 
where $\symbolflip$ and $\symbolremain$ are parameters controlling the noise level in \technique's smoothing distribution $\mu$. 
\end{theorem}

\begin{theorem}[Compute $\cleandistmass{\subspace_{c, t}}$]
\label{thm:selectflip_main2}
\begin{align*}
   \cleandistmass{\subspace_{c, t}} = \underset{\outcome \sim \cleandist}{\PP}(\outcome \text{ satisfies Eq~\ref{eq:Ldef1}}) \underset{\outcome \sim \cleandist}{\PP}(\outcome \text{ satisfies Eq~\ref{eq:Ldef3}} \mid \outcome \text{ satisfies Eq~\ref{eq:Ldef1}}),
\end{align*}
where $\PP_{\outcome \sim \cleandist}(\outcome \text{ satisfies Eq~\ref{eq:Ldef1}}) = \mathrm{Binom}(c;k,\frac{\poisonins}{n})$ is the PMF of the binomial distribution and 
\begin{small}
\begin{align}
    T(c,t) \triangleq \underset{\outcome \sim \cleandist}{\PP}(\outcome \text{ satisfies Eq~\ref{eq:Ldef3}} \mid \outcome \text{ satisfies Eq~\ref{eq:Ldef1}}) = \sum_{\substack{0 \le \diffu_i, \diffv_i \le d, \forall i\in[0,d] \\ \diffu_0-\diffv_0 + \ldots + \diffu_c-\diffv_c = t}} \prod_{i=0}^c L(\diffu_i,\diffv_i;\poisonfeat, d)\symbolflip^{\diffu_i}\symbolremain^{d-\diffu_i},\label{eq:prd'}
\end{align}
\end{small}%
where $L(\diffu,\diffv;\poisonfeat, d)$ is the same quantity defined in \citet{tightflipbound}.
\end{theorem}


\begin{remark}
We can compute $\poisondistmass{\subspace_{c, t}}$ as $\eta_{c,t}\cleandistmass{\subspace_{c, t}}$ by the definition of $\eta_{c,t}$.
\end{remark}

\textbf{Efficient algorithm to compute Eq~\ref{eq:prd'}.}
The following algorithm computes $T(c,t)$ efficiently in time~$O(d^3 + k^2d^2)$: 1) Computing $L(u, v;\poisonfeat, d)$ takes $O(d^3)$ (see \Cref{sec: proofs} for details). 2) Computing $T(0,t)$ by the definition in Eq~\ref{eq:prd'} takes $O(kd^2)$. 3) Computing $T(c,t)$ for $c\ge 1$ by the following equation takes $O(k^2d^2)$.
\begin{small}
    \begin{align}
        T(c, t) = \sum_{t_1=\max(-d, t-cd)}^{\min(d, t+cd)} T(c-1,t - t_1)T(0,t_1) \label{eq:efficient_tct}
    \end{align}
\end{small}%

\begin{theorem}[Correctness of the Algorithm]
\label{thm:selectflip_tct}
$T(c,t)$ in Eq~\ref{eq:efficient_tct} is the same as the one in Eq~\ref{eq:prd'}.
\end{theorem}

The above computation still applies to perturbation function $\featfliplabelflip$ and $\labelflip$. 
Intuitively, flipping the label can be seen as flipping another dimension in the input features (Details in \Cref{sec: other_perturbs}).

\textbf{Practical perspective.} For each test input $\bfxtest$, we need to estimate $\firstp$ for the smoothed algorithm $\salg$ given the benign dataset $D$.
We use Monte Carlo sampling to compute $\firstp$ by the Clopper-Pearson interval.
We also reuse the trained algorithms for each test input in the test set by using \textit{Bonferroni correction}.
We memoize the certified radius by enumerating all possible $\firstp$ beforehand so that checking Eq~\ref{eq:certgoal1} can be done in constant time in an online scenario (details in \Cref{sec: appendix_practical}).
We cannot use floating point numbers because some intermediate values are too small to store in the floating point number format. 
Instead, we use rational numbers which represent the nominator and denominator in large numbers.

\section{An Efficient Relaxation for Computing a Certified Radius}
\label{sec:imprecise}
Theorem~\ref{thm:np_bound} requires computing the likelihood ratio of a large number of subspaces $\subspace_1,\ldots,\subspace_m$.
We propose a generalization of the Neyman--Pearson lemma that \textbf{underapproximates} the subspaces by a small subset and computes a lower bound $\lb_\delta$ for $\lb$.
We then show how to choose a subset of subspaces that yields a tight underapproximation. 


\textbf{A relaxation of the Neyman--Pearson lemma.}
The key idea is to use Theorem~\ref{thm:np_bound} to examine only a \textit{subset} of the subspaces $\{\subspace_i\}_i$.
Suppose that we partition the subspaces $\{\subspace_i\}_i$ into two sets $\{\subspace_i\}_{i\in \indexset}$
and $\{\subspace_i\}_{i\not\in \indexset}$, using 
a set of indices $\indexset$.
We define a new lower bound $\lb_\delta$ by applying Theorem~\ref{thm:np_bound} to the first group $\{\subspace_i\}_{i\in \indexset}$ and underapproximating $\firstp$ in Theorem~\ref{thm:np_bound} as $\firstp - \sum_{i\notin \indexset} \cleandistmass{\subspace_i}$.
Intuitively, the underapproximation ensures that any subspace in $\{\subspace_i\}_{i\not\in \indexset}$ will not contribute to the lower bound, even though these subspaces may have contributed to the precise $\lb$ computed by Theorem~\ref{thm:np_bound}.

The next theorem shows that $\lb_\delta$ will be smaller than $\lb$ by an error term $\delta$ that is a function of the partition $\indexset$.
The gain is in the number of subspaces we have to consider, which is now $|B|$.

\begin{theorem}[Relaxation of the Lemma]
\label{thm:np_non_tight}
Define $\lb$ for the subspaces $\subspace_1, \ldots, \subspace_m$ as in \cref{thm:np_bound}.
Define $\lb_\delta$ for $\{\subspace_i\}_{i\in \indexset}$ as in \cref{thm:np_bound}  
by underapproximating $\firstp$ as 
$\firstp - \sum_{i\notin \indexset} \cleandistmass{\subspace_i}$, then we have 
\textbf{Soundness:} $\lb_\delta \le \lb$ and \textbf{$\delta$-Tightness:} Let $\delta \triangleq \sum_{i \notin \indexset} \poisondistmass{\subspace_i}$, then $\lb_\delta + \delta \ge \lb$.
\end{theorem}

\begin{example}
\label{exp: relaxation}
Consider $\subspace_1, \subspace_2$, and $\subspace_3$ from \Cref{exp: lemma}. If we set $\indexset = \{1, 2\}$ and underapproximate $\firstp$ as $\firstp - \cleandistmass{\subspace_3} = 0.85$,
then we have $\lb_\delta = 0.1 + (0.85 - 0.4) / 1 = 0.55$, which can still certify Eq.~\ref{eq:certgoal1}.
However, $\lb_\delta$ is not close to the original $\lb = 0.8$ because the error $\delta = 0.4$ is large in this example.
Next, we will show how to choose $\indexset$ as small as possible while still keeping $\delta$ small. 
\end{example}

\textbf{Speeding up radius computation.}
The total time for computing $\lb$ consists of two parts, (1) the efficient algorithm~(Eq~\ref{eq:efficient_tct}) computes $T(c,t)$ in $O(d^3+k^2d^2)$, and (2) Theorem~\ref{thm:np_bound} takes $O(k^2d)$ for a given $\firstp$ because there are $O(k^2d)$ subspaces. 
Even though we have made the computation polynomial by the above two techniques in \cref{sec: featflip}, it can still be slow for large bag sizes $k$, which can easily be in hundreds or thousands.

We will apply Theorem~\ref{thm:np_non_tight} to replace the bag size $k$ with a small constant $\smallk$.
Observe that $\cleandistmass{\subspace_{c,t}}$ and $\poisondistmass{\subspace_{c,t}}$, as defined in \cref{sec: featflip}, are negligible for large $c$, i.e., the number of perturbed indices in the smoothed dataset.
Intuitively, if only a small portion of the training set is perturbed, it is unlikely that we select a large number of perturbed indices in the smoothed dataset.
We underapproximate the full subspaces to $\{\subspace_{c,t}\}_{(c,t) \in \indexset}$ by choosing the set of indices $\indexset=\{(c,t) \mid c\le \smallk , |t| \le (c+1)d\}$, where $\smallk$ is a constant controlling the error $\delta$, which can be computed as $\delta = \sum_{i \notin \indexset}\poisondistmass{\subspace_i} = 1 - \sum_{c=0}^\smallk \mathrm{Binom}(c;k,\frac{\poisonins}{n})$.
Theorem~\ref{thm:np_non_tight} reduces the number of subspaces to $|\indexset| = O(\smallk^2d)$.
As a result, all the $k$ appearing in previous time complexity can be replaced with $\smallk$.
\begin{example}
Suppose $k=150$ and $\frac{r}{n}=0.5\%$, choosing $\smallk = 6$ leads to $\delta = 1.23\times 10^{-5}$. 
In other words, it speeds up almost 625X for computing $T(c,t)$ and for applying Theorem~\ref{thm:np_bound}.
\end{example}

\section{Experiments}
\label{sec: experiment}
The implementation of \technique is publicly available\footnote{\url{https://github.com/ForeverZyh/defend_framework}}.
In this section, we evaluate \technique against trigger-less and backdoor attacks and compare \technique with existing work.
Note that we apply the relaxation in Section~\ref{sec:imprecise} to \technique in all experiments and set $\delta = 10^{-4}$.

\textbf{Datasets.}
We conduct experiments on MNIST, CIFAR10, EMBER~\cite{ember}, and Contagio (\url{http://contagiodump.blogspot.com}).
MNIST and CIFAR10 are datasets for image classification. EMBER and Contagio are datasets for malware detection,
where data poisoning can lead to disastrous security consequences.
To align with existing work, we select subsets of MNIST and CIFAR10 as MNIST-17, MNIST-01, and CIFAR10-02 in some experiments, e.g., MNIST-17 is a subset of MNIST with classes 1 and 7. 
We discretize all the features when applying \technique.
We use clean datasets except in the comparison with RAB~\cite{RAB}, where we follow their experimental setup and use backdoored datasets generated by BadNets~\cite{badnets}.
A detailed description of the datasets is in \Cref{sec: appendix_data}.

\textbf{Models.} We train neural networks for MNIST, CIFAR10, EMBER and random forests for Contagio. 
Unless we specifically mention the difference in some experiments, 
whenever we compare \technique to an existing work, we will use the same network structures, hyper-parameters, and data augmentation as the compared work, and we train $N=1000$ models and set the confidence level as $0.999$ for each configuration.
All the configurations we used can be found in the implementation. 

\textbf{Metrics.}
For each test input $(\bfxtest_i, y_i)$, algorithm $\salg$ will predict a label and the certified radius $\poisonins_i$.
Assuming that the attacker had poisoned $\assumeradii\%$ of the examples in the training set, we define \textbf{certified accuracy} as the percentage of test inputs that are correctly classified and their certified radii are no less than $\assumeradii$, i.e., $\sum_{i=1}^m \mathds{1}_{\smodelpred{D}{\bfxtest_i} = y_i \wedge \frac{\poisonins_i}{n} \ge \assumeradii\%}$.
We define \textbf{normal accuracy} as the percentage of test inputs that are correctly classified, i.e., $\sum_{i=1}^m \mathds{1}_{\smodelpred{D}{\bfxtest_i} = y_i}$.

\subsection{Defending Against Trigger-less Attacks}
Two model-agnostic certified approaches, Bagging~\cite{bagging} and LabelFlip~\cite{labelflip}, can defend against trigger-less attacks.
\technique outperforms LabelFlip (comparison in~\Cref{sec: appendix_res}).
We evaluate \technique on the perturbation $\featflip$ using MNIST, CIFAR10, EMBER, and Contagio and compare \technique to Bagging. We provide comparisons with Bagging on other perturbation spaces in~\Cref{sec: appendix_res}.

We present \technique with different noise levels (different probabilities of $\symbolremain$ when flipping), denoted as \technique-$\symbolremain$.
When comparing to Bagging, we use the same $k$, the size of sampled datasets, for a fair comparison.
Furthermore, we tune the parameter $k$ in Bagging to match the normal accuracy with the \technique-$\symbolremain$ setting, and denote this setting as Bagging-$\symbolremain$.
Concretely, we set $k=100, 1000, 300, 30$ for MNIST, CIFAR10, EMBER, and Contagio respectively when comparing to Bagging. 
We tune $k=80, 280$ for Bagging-0.9 on MNIST and Bagging-0.95 on EMBER, respectively.
And we set $k=50$ for MNIST when comparing to LabelFlip.

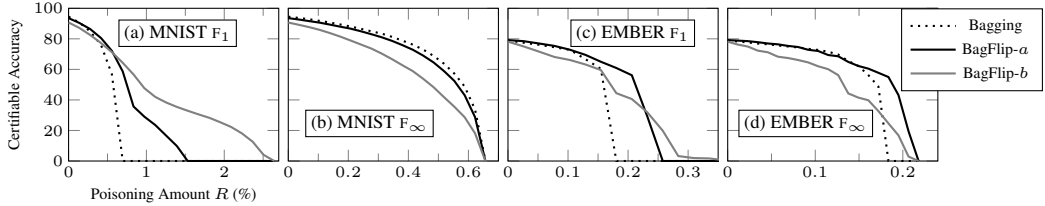
\begin{figure}[t!]
\centering
\tiny
\pgfplotsset{filter discard warning=false}
\pgfplotsset{every axis legend/.append style={
at={(1.5,1)},
anchor=north east}} 

\pgfplotscreateplotcyclelist{whatever}{%
    black,thick,dotted,every mark/.append style={fill=blue!80!black},mark=none\\%
    black,thick,every mark/.append style={fill=red!80!black},mark=none\\%
    gray,thick,every mark/.append style={fill=blue!80!black},mark=none\\%
    }
    
\begin{tikzpicture}
    \begin{groupplot}[
            group style={
                group size=4 by 1,
                horizontal sep=.05in,
                vertical sep=.05in,
                ylabels at=edge left,
                yticklabels at=edge left,
                xlabels at=edge bottom,
                xticklabels at=edge bottom,
            },
            height=.8in,
            xlabel near ticks,
            ylabel near ticks,
            scale only axis,
            width=0.2*\textwidth,
            xmin=0,
            ymin=0,
            ymax=100,
        ]

        \nextgroupplot[
            ylabel=Certifiable Accuracy,
            xmax=2.7,
            xtick={0,1,2},
            minor xtick={0.5,1.5,2.5},
            cycle list name=whatever,
            xlabel=Poisoning Amount $\assumeradii$ (\%)]
        \addplot table [x=x,y=0, col sep=comma]{data/trigger_less/mnist_1.csv};
        \addplot table [x=x,y=1, col sep=comma]{data/trigger_less/mnist_1.csv};
        \addplot table [x=x,y=2, col sep=comma]{data/trigger_less/mnist_1.csv};
        
        \nextgroupplot[
            xmax=0.7,
            xtick={0,0.2,0.4,0.6},
            minor xtick={0.1,0.3,0.5,0.7},
             cycle list name=whatever,
        ]
        \addplot table [x=x,y=0, col sep=comma]{data/trigger_less/mnist_inf.csv};
        \addplot table [x=x,y=1, col sep=comma]{data/trigger_less/mnist_inf.csv};
        \addplot table [x=x,y=2, col sep=comma]{data/trigger_less/mnist_inf.csv};

        \nextgroupplot[
        xmax=0.35,
            xtick={0,0.1,0.2,0.3},
            minor xtick={0.05,0.15,0.25},
             cycle list name=whatever,
        ]
        \addplot table [x=x,y=0, col sep=comma]{data/trigger_less/ember_1.csv};
        \addplot table [x=x,y=1, col sep=comma]{data/trigger_less/ember_1.csv};
        \addplot table [x=x,y=2, col sep=comma]{data/trigger_less/ember_1.csv};
        
        \nextgroupplot[
        xmax=0.24,
            xtick={0,0.1,0.2},
            minor xtick={0.05,0.15},
             cycle list name=whatever,
        ]
        \addplot table [x=x,y=0, col sep=comma]{data/trigger_less/ember_inf.csv};
        \addplot table [x=x,y=1, col sep=comma]{data/trigger_less/ember_inf.csv};
        \addplot table [x=x,y=2, col sep=comma]{data/trigger_less/ember_inf.csv};

\legend{Bagging, \technique-$a$, \technique-$b$};

    
    \end{groupplot}
    \node[draw] at (1.45,1.7) {\scriptsize (a) MNIST $\featflipone$};
    \node[draw] at (4,0.5) {\scriptsize (b) MNIST $\featflipinf$};
    \node[draw] at (7.5,1.7) {\scriptsize (c) EMBER $\featflipone$};
    \node[draw] at (9.8,0.5) {\scriptsize (d) EMBER $\featflipinf$};

\end{tikzpicture}
\vspace{-1em}
\caption{Comparison to Bagging on MNIST and EMBER, showing the certified accuracy at different poisoning amounts $\assumeradii$. 
For MNIST: $a=0.9$ and $b=0.8$. For EMBER: $a=0.95$ and $b=0.9$.
}
\vspace{-2em}
\label{fig:compare_bagging}
\end{figure}

\begin{table}
\centering
\caption{Certified accuracy on MNIST and EMBER with perturbations $\featflipone$ and $\featflipinf$. Note that the certified accuracies are the same poisoning amount $R=0$  because we reuse the trained models. }
\setlength{\tabcolsep}{3pt}
\begin{small}
\begin{tabular}{llllllllllllll}
\toprule
            && \multicolumn{6}{c}{$\featflipone$}                 & \multicolumn{6}{c}{$\featflipinf$}        \\ 
\cmidrule(lr){3-8} \cmidrule(lr){9-14}
\multirow{4}{*}{\rotatebox[origin=c]{90}{MNIST}} &$\assumeradii$   & $0$ & $0.5$ & $1.0$ & $1.5$ & $2.0$ & $2.5$ & $0$ & $0.13$ & $0.27$ & $0.40$ & $0.53$ & $0.83$  \\
\cmidrule{2-14}
 & Bagging     & \textbf{94.54}           & 66.84& 0    & 0    & 0     & 0     & \textbf{94.54} & \textbf{90.83} &	\textbf{85.45} &	77.61 &	61.46 & 0     \\
 & Bagging-0.9     & 93.58     & 71.11     & 0         & 0        & 0   & 0    & 93.58     & 89.80 & 84.92 & \textbf{78.60} & \textbf{68.11} & 0 \\
& \technique-0.9 & 93.62           & \textbf{75.95} & 27.73& 4.02 & 0     & 0     & 93.62 & 89.45            & 83.62& 74.19& 56.65& 0     \\
& \technique-0.8 & 90.72           & 73.94& \textbf{46.20} & \textbf{33.39} & \textbf{24.23} & \textbf{5.07}  & 90.72 & 84.11            & 74.50& 60.56& 39.21& 0    \\
\midrule
\multirow{4}{*}{\rotatebox[origin=c]{90}{EMBER}} &$\assumeradii$   & $0$ & $0.07$ & $0.13$ & $0.20$ & $0.27$ & $0.33$ & $0$ & $0.05$ & $0.10$ & $0.15$ & $0.20$ & $0.25$  \\
\cmidrule{2-14}
 & Bagging & \textbf{82.65} & 75.11 & 66.01 & 0 & 0 & 0
& \textbf{82.65} & 76.30 & 72.94 & 61.78 & 0 & 0\\
& Bagging-0.95      & 79.06     & 75.32     & \textbf{70.19} & 14.74     & 0    & 0    & 79.06     & 76.23     &  \textbf{73.50} & \textbf{68.45} & 14.74     & 0     \\
& \technique-0.95 & 79.17 & \textbf{75.93} & 69.30 & \textbf{57.36} & 0 & 0
& 79.17 & \textbf{76.83} & 72.41 & 62.04 & \textbf{30.36} & 0\\
& \technique-0.9 & 78.18 & 69.40 & 62.11 & 41.21 & \textbf{13.89} & \textbf{1.70} 
& 78.18 & 70.79 & 63.16 & 41.24 & 11.64 & 0\\
\bottomrule
\end{tabular}
\end{small}
\label{tab:trigger_less_res}
\vspace{-1em}
\end{table}

\textbf{Comparison to Bagging.} 
Bagging on a discrete dataset is a special case of \technique when $\symbolremain=1$, i.e., no noise is added to the dataset. (We present the results of bagging on the original dataset (undiscretized) in 
\Cref{sec: appendix_res}).
Table~\ref{tab:trigger_less_res} and Fig~\ref{fig:compare_bagging} show the results of \technique and Bagging on perturbations $\featflipone$ and $\featflipinf$ over MNIST and EMBER. 
The results of CIFAR10 and Contagio are similar and shown in \Cref{sec: appendix_res}.
$\featflipone$ only allows the attacker to modify one feature in each training example, and $\featflipinf$ allows the attacker to modify each training example arbitrarily without constraint.
For $\featflipone$ on MNIST, \technique-0.9 performs better than Bagging after $\assumeradii = 0.19$ and \technique-0.8 still retains non-zero certified accuracy at $\assumeradii=2.5$ while Bagging's certified accuracy drops to zero after $\assumeradii=0.66$.
We observe similar results on EMBER for $\featflipone$ in Table~\ref{tab:trigger_less_res}, except $\assumeradii=0.13$ when compared to Bagging-0.95.
We argue that it is possible to tune a best combination of $k$ and $\symbolremain$ for \technique, like we tune $k$ for Bagging-0.95, and achieve a better result while maintaining similar normal accuracy. 
However, we do not conduct hyperparameter-tuning for \technique because of its computation cost.
\textbf{\technique achieves higher certified accuracy than Bagging when the poisoning amount is large for $\featflipone$.}

For $\featflipinf$ on MNIST, Bagging performs better than \technique across all $\assumeradii$ because the noise added by \technique to the training set hurts the accuracy. 
However, we find that the noise added by \technique helps it perform better for $\featflipinf$ on EMBER.
Specifically, \technique achieves similar certified accuracy as Bagging at small radii and \technique performs better than Bagging after $\assumeradii=0.15$.
\textbf{Bagging achieves higher certified accuracy than \technique for $\featflipinf$. Except in EMBER, \technique achieves higher certified accuracy than Bagging when the amount of poisoning is large.}

  

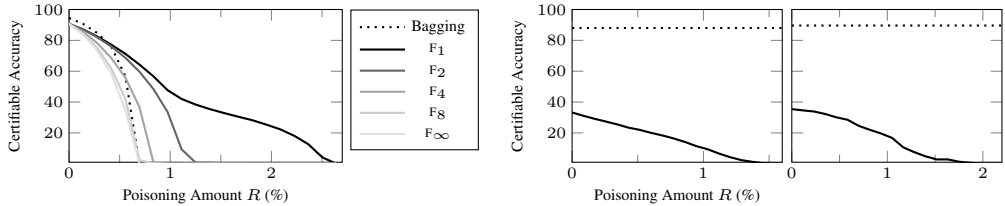
\begin{figure}[t]
    \centering\tiny
    \subfigure{\begin{tikzpicture}
\pgfplotsset{filter discard warning=false}
\pgfplotsset{every axis legend/.append style={
at={(1.5,1)},
anchor=north east}} 

\pgfplotscreateplotcyclelist{whatever}{%
    black,thick,dotted,every mark/.append style={fill=blue!80!black},mark=none\\%
    black,thick,every mark/.append style={fill=red!80!black},mark=none\\%
    black!60,thick,every mark/.append style={fill=blue!80!black},mark=none\\%
    black!36,thick,every mark/.append style={fill=blue!80!black},mark=none\\%
    black!20,thick,every mark/.append style={fill=blue!80!black},mark=none\\%
    black!12,thick,every mark/.append style={fill=blue!80!black},mark=none\\%
    }
    \begin{groupplot}[
            group style={
                group size=1 by 1,
                horizontal sep=.05in,
                vertical sep=.05in,
                ylabels at=edge left,
                yticklabels at=edge left,
                xlabels at=edge bottom,
                xticklabels at=edge bottom,
            },
            height=.8in,
            xlabel near ticks,
            ylabel near ticks,
            scale only axis,
            width=0.26*\textwidth,
            xmin=0,
            ymin=1,
            ymax=100,
        ]

        \nextgroupplot[
            ylabel=Certifiable Accuracy,
            xmax=2.7,
            xtick={0,1,2},
            minor xtick={0.5,1.5,2.5},
            cycle list name=whatever,
            xlabel=Poisoning Amount $\assumeradii$ (\%)]
        \addplot table [x=x,y=0, col sep=comma]{data/trigger_less/mnist_0.8_s.csv};
        \addplot table [x=x,y=1, col sep=comma]{data/trigger_less/mnist_0.8_s.csv};
        \addplot table [x=x,y=2, col sep=comma]{data/trigger_less/mnist_0.8_s.csv};
        \addplot table [x=x,y=3, col sep=comma]{data/trigger_less/mnist_0.8_s.csv};
         \addplot table [x=x,y=4, col sep=comma]{data/trigger_less/mnist_0.8_s.csv};
          \addplot table [x=x,y=6, col sep=comma]{data/trigger_less/mnist_0.8_s.csv};

\legend{Bagging, $\featflipone$, $\featfliptwo$,$\featflipfour$, $\featflipeight$, $\featflipinf$};

    
    \end{groupplot}

\end{tikzpicture}\label{fig:mnist_0.8_s}}
    \hspace{1em}
    \subfigure{    
\begin{tikzpicture}
\pgfplotsset{filter discard warning=false}
\pgfplotsset{every axis legend/.append style={
at={(1.5,1)},
anchor=north east}} 

\pgfplotscreateplotcyclelist{whatever}{%
    black,thick,every mark/.append style={fill=blue!80!black},mark=none\\%
    black,thick,dotted,every mark/.append style={fill=red!80!black},mark=none\\%
    }
    \begin{groupplot}[
            group style={
                group size=2 by 1,
                horizontal sep=.05in,
                vertical sep=.05in,
                ylabels at=edge left,
                yticklabels at=edge left,
                xlabels at=edge bottom,
                xticklabels at=edge bottom,
            },
            height=.8in,
            xlabel near ticks,
            ylabel near ticks,
            scale only axis,
            width=0.2*\textwidth,
            xmin=0,
            ymin=1,
            ymax=100,
        ]

        \nextgroupplot[
            ylabel=Certifiable Accuracy,
            xmax=1.6,
            xtick={0,1},
            minor xtick={0.5,1.5},
            cycle list name=whatever,
            xlabel=Poisoning Amount $\assumeradii$ (\%)]
        \addplot table [x=x,y=1, col sep=comma]{data/backdoor/bd_mnist.csv};
        \addplot table [x=x,y=1-normal, col sep=comma]{data/backdoor/bd_mnist.csv};
        
        \nextgroupplot[
            xmax=2.2,
            xtick={0,1,2},
            minor xtick={0.5,1.5},
            cycle list name=whatever]
        \addplot table [x=x,y=0, col sep=comma]{data/backdoor/bd_contagio.csv};
        \addplot table [x=x,y=0-normal, col sep=comma]{data/backdoor/bd_contagio.csv};


    
    \end{groupplot}

\end{tikzpicture}\label{fig:bd_mnist_contagio}}
    \vspace{-1em}
    \caption{(a) \technique-0.8 on MNIST against $\featflip$ with different $\poisonfeat$. $\poisonfeat=8$ almost overlaps with $\poisonfeat=\infty$. (b) \technique-0.8 on MNIST and \technique-0.9 on Contagio against backdoor attack with $\featflipone$. Dashed lines show normal accuracy.}
    \vspace{-2em}
\end{figure}

We also study the effect of different $\poisonfeat$ in the perturbation function $\featflip$. 
Figure~\ref{fig:mnist_0.8_s} shows the result of \technique-0.8 on MNIST.
\technique has the highest certified accuracy for $\featflipone$. 
As $\poisonfeat$ increases, the result monotonically converges to the curve of \technique-0.8 in Figure~\ref{fig:compare_bagging}(b).
Bagging neglects the perturbation function and performs the same across all $\poisonfeat$.  
Bagging performs better than \technique-0.8 when $\poisonfeat\geq 8$.
Other results for different datasets and different noise levels follow a similar trend (see \Cref{sec: appendix_res}).
\textbf{\technique performs best at $\featflipone$ and monotonically degenerates to $\featflipinf$ as $\poisonfeat$ increases.}

\subsection{Defending Against Backdoor Attacks}

Two model-agnostic certified approaches, FeatFlip~\cite{featureflip} and RAB~\cite{RAB}, can defend against backdoor attacks.
We compare \technique to FeatFlip on MNIST-17 perturbed using $\featfliplabelflipone$, and compare \technique to RAB on poisoned MNIST-01 and CIFAR10-02 (by BadNets~\cite{badnets}) perturbed using $\featflipone$.
We further evaluate \technique on the perturbation $\featflipone$ using MNIST and Contagio, on which existing work is unable to compute a meaningful certified radius.



\begin{wraptable}{R}{0.5\textwidth}
\vspace{-1.4em}
\centering
\caption{Compared to FeatFlip on $\featfliplabelflipone$ and RAB $\featflipone$ with normal accuracy~(Acc.) and certified accuracy at different poisoning amount $\assumeradii$~(CF@$\assumeradii$).
}
\setlength{\tabcolsep}{3pt}
\begin{small}
\begin{tabular}{llllll}
\toprule
\multirow{4}{*}{\rotatebox[origin=c]{90}{\scriptsize MNIST-17}} && Acc. & CF@0 & CF@0.5 & CF@1.0         \\
\cmidrule{3-6}
 &
 FeatFlip  & 98 & 36 & 0 & 0 \\
 & \technique-0.95 & 97 & \textbf{81} & \textbf{60} & 0 \\
 & \technique-0.9 & 97 & 72 & 46 & \textbf{4}\\
\midrule
 \multirow{4}{*}{\rotatebox[origin=c]{90}{\scriptsize MNIST-01}} && Acc. & CF@0 & CF@1.5 & CF@3.0         \\
 \cmidrule{3-6}
 & RAB & 100.0 & 74.4 & 0 & 0\\
& \technique-0.8 & 99.6 & \textbf{98.4} & \textbf{96.5} & 84.6 \\
& \technique-0.7 & 99.5 & 98.0 & 95.8 & \textbf{91.9}\\
\midrule
\multirow{4}{*}{\rotatebox[origin=c]{90}{\scriptsize CIFAR10-02}}  && Acc. & CF@0 & CF@0.1 & CF@0.2         \\
\cmidrule{3-6}
 & RAB & 73.3 & 0 & 0 & 0\\
& \technique-0.8 & 73.1 & 16.8 & 11.0 & 6.8 \\
& \technique-0.7 & 71.7 & \textbf{41.0} & \textbf{34.6} & \textbf{29.2} \\
\bottomrule
\end{tabular}
\end{small}

\label{tab:backdoor_res}
\vspace{-2em}
\end{wraptable}

\textbf{Comparison to FeatFlip.} FeatFlip scales poorly compared to \technique because its computation of the certified radius is polynomial in the size of the training set. \technique's computation is polynomial in the size of the bag instead of the size of the whole training set, and it uses a relaxation of the Neyman--Pearson lemma for further speed up.
\textbf{\technique is more scalable than FeatFlip.}

We directly cite FeatFlip's results from~\citet{featureflip} and note that FeatFlip is evaluated on a subset of MNIST-17.
As shown in Table~\ref{tab:backdoor_res}, \technique achieves similar normal accuracy, but much higher certified accuracy across all $\assumeradii$~(see full results in \Cref{sec: appendix_res}) than FeatFlip.
\textbf{\technique significantly outperforms FeatFlip against $\featfliplabelflipone$ on MNIST-17.}

\textbf{Comparison to RAB.} RAB assumes a perturbation function that perturbs the input within an $l_2$-norm ball of radius $\poisonfeat$. We compare \technique to RAB on $\featflipone$, where the $l_0$-norm ball (our perturbation function) and $l_2$-norm ball are the same because the feature values are within $[0,1]$.
Since RAB targets single-test-input prediction, we do not use Bonferroni correction for \technique as a fair comparison.
We directly cite RAB's results from~\citet{RAB}.
Table~\ref{tab:backdoor_res} shows that on MNIST-01 and CIFAR10-02, \technique achieves similar normal accuracy, but a much higher certified accuracy than RAB for all values of $\assumeradii$~(detailed figures in \Cref{sec: appendix_res}).
\textbf{\technique significantly outperforms RAB against $\featflipone$ on MNIST-01 and CIFAR10-02.}

\textbf{Results on MNIST and Contagio.}
Figure~\ref{fig:bd_mnist_contagio} shows the results of \technique on MNIST and Contagio. 
When fixing $\assumeradii$, the certified accuracy for the backdoor attack is much smaller than the certified accuracy for the trigger-less attack~(Figure~\ref{fig:compare_bagging} and Figure~\ref{fig:compare_bagging_full} in \Cref{sec: appendix_res}) because backdoor attacks are strictly stronger than trigger-less attacks.
\technique cannot provide effective certificates for backdoor attacks on the more complex datasets CIFAR10 and EMBER, i.e., the certified radius is almost zero.
\textbf{\technique can provide certificates against backdoor attacks on MNIST and Contagio, while \technique's certificates are not effective for CIFAR10 and EMBER.}


\subsection{Computation Cost Analysis}
We discuss the computation cost of \technique on the MNIST dataset and compare to other baselines.

\textbf{Training.} The cost of \technique during training is similar to all the baselines because \technique only adds noise in the training data. 
\technique and other baselines take about 16 hours on a single GPU to train $N=1000$ classifiers on the MNIST dataset.

\textbf{Inference.} At inference time, \technique first evaluates the predictions of $N$ classifiers, and counts how many classifiers have the majority label ($N_1$) and how many have the runner-up label ($N_2$). 
Then, \technique uses a prepared lookup table to query the radius certified by $N_1$ and $N_2$.
The inference time for each example contains the evaluation of $N$ classifiers and an O(1) table lookup. Hence, there is no difference between \technique and other baselines.

\textbf{Preparation.} \technique needs to prepare a table of size O($N^2$) to perform efficient lookup at  inference time. 
The time complexity of preparing each table entry is presented in Sections~\ref{sec:precise_approach} and ~\ref{sec:imprecise}.
On the MNIST dataset, \technique with the relaxation proposed in Section~\ref{sec:imprecise} ($\delta = 10^{-4}$) needs 2 hours to prepare the lookup table on a single core. 
However, the precise \technique proposed in Section~\ref{sec:precise_approach} needs 85 hours to prepare the lookup table. 
Bagging also uses a lookup table that can be built in 16 seconds on MNIST (Bagging only needs to do a binary search for each entry). 
FeatFlip needs approximately 8000 TB of memory to compute its table. Thus, FeatFlip is infeasible to run on the full MNIST dataset. FeatFlip is only evaluated on a subset of the MNIST-17 dataset containing only 100 training examples.
RAB does not need to compute the lookup table because it has a closed-form solution for computing the certified radius.

\textbf{\technique has similar training and inference time compared to other baselines. The relaxation technique in Section~\ref{sec:imprecise} is useful to reduce the preparation time from 85 hours to 2 hours.} 
Even with the relaxation, \technique needs more preparation time than Bagging and RAB. We argue that the preparation of \technique is feasible because it only takes 12.5\% of the time required by training.
\section{Conclusion, Limitations, and Future Work}
\label{sec: conclusion}
We presented \technique, a certified probabilistic approach that can effectively defend against both trigger-less and backdoor attacks.
We foresee many future improvements to \technique. First, \technique treats both the data and the underlying machine learning models as closed boxes. Assuming a specific data distribution and training algorithm can further improve the computed certified radius. 
Second, \technique uniformly flips the features and the label, while it is desirable to adjust the noise levels for the label and important features for better normal accuracy according to the distribution of the data. 
Third, while probabilistic approaches need to retrain thousands of models after a fixed number of predictions, the deterministic approaches can reuse models for every prediction. Thus, it is interesting to develop a deterministic model-agnostic approach that can defend against both trigger-less and backdoor attacks.

\section*{Acknowledgments and Disclosure of Funding}
We thank the anonymous reviewers and Anna P. Meyer for their thoughtful and generous comments and Gurindar S. Sohi for giving us access to his GPUs. This work is supported by the CCF-1750965, CCF-1918211, CCF-1652140, CCF-2106707, CCF-1652140, a Microsoft Faculty Fellowship, and gifts and awards from Facebook and Amazon.


\bibliography{main}
\bibliographystyle{plainnat}


\section*{Checklist}


\begin{enumerate}

\item For all authors...
\begin{enumerate}
  \item Do the main claims made in the abstract and introduction accurately reflect the paper's contributions and scope?
    \answerYes{}
  \item Did you describe the limitations of your work?
    \answerYes{See Section~\ref{sec: conclusion}.}
  \item Did you discuss any potential negative societal impacts of your work?
    \answerNA{}
  \item Have you read the ethics review guidelines and ensured that your paper conforms to them?
    \answerYes{}
\end{enumerate}

\item If you are including theoretical results...
\begin{enumerate}
  \item Did you state the full set of assumptions of all theoretical results?
    \answerYes{}
        \item Did you include complete proofs of all theoretical results?
    \answerYes{We provide them in supplementary materials.}
\end{enumerate}

\item If you ran experiments...
\begin{enumerate}
  \item Did you include the code, data, and instructions needed to reproduce the main experimental results (either in the supplemental material or as a URL)?
    \answerYes{}
  \item Did you specify all the training details (e.g., data splits, hyperparameters, how they were chosen)?
    \answerYes{}
        \item Did you report error bars (e.g., with respect to the random seed after running experiments multiple times)?
    \answerNo{As our results are aggregated from thousands of models with the same statistical confidence level, we do not report the error bars.}
        \item Did you include the total amount of compute and the type of resources used (e.g., type of GPUs, internal cluster, or cloud provider)?
    \answerYes{}
\end{enumerate}

\item If you are using existing assets (e.g., code, data, models) or curating/releasing new assets...
\begin{enumerate}
  \item If your work uses existing assets, did you cite the creators?
    \answerYes{}
  \item Did you mention the license of the assets?
    \answerYes{MIT license}
  \item Did you include any new assets either in the supplemental material or as a URL?
    \answerYes{}
  \item Did you discuss whether and how consent was obtained from people whose data you're using/curating?
    \answerNA{They are either under MIT license or allowed for public usage.}
  \item Did you discuss whether the data you are using/curating contains personally identifiable information or offensive content?
    \answerNA{There is no personally identifiable information or offensive content.}
\end{enumerate}

\item If you used crowdsourcing or conducted research with human subjects...
\begin{enumerate}
  \item Did you include the full text of instructions given to participants and screenshots, if applicable?
    \answerNA{}
  \item Did you describe any potential participant risks, with links to Institutional Review Board (IRB) approvals, if applicable?
    \answerNA{}
  \item Did you include the estimated hourly wage paid to participants and the total amount spent on participant compensation?
    \answerNA{}
\end{enumerate}

\end{enumerate}
\clearpage
\appendix
\section{Probability Mass Functions of Distributions for $\featfliplabelflip$, $\featflip$, and $\labelflip$}
\label{sec: pmfs}
\paragraph{Distribution $\selectflip$ for $\featflip$.} 
The distribution $\selectflip(D, \bfxtest)$ describes the outcomes of $\sample{D}$ and $\sample{\bfxtest}$.
Each outcome $\sample{D}$ and $\sample{\bfxtest}$ can be represented as a combination of 1) selected indices $w_1, \ldots, w_k$, 2) smoothed training examples $\bfx_{w_1}', \ldots, \bfx_{w_k}'$, and 3) smoothed test input $\bfx'$.
The probability mass function (PMF) of $\selectflip(D, \bfxtest)$ is,
\begin{align}
    p_{\selectflip(D,\bfxtest)}(\sample{D}, \sample{\bfxtest}) = \frac{\symbolremain^{(k+1)d}}{n^k}  \left(\frac{\symbolflip}{\symbolremain}\right)^{\sum_{i=1}^k \dis{\bfx_{w_i}'}{\bfx_{w_i}} + \dis{\bfxtest'}{\bfxtest}},\label{eq:jointdist}
\end{align}
where $d$ is the dimension of the input feature, $\symbolflip=\frac{1-\symbolremain}{K}$, $K+1$ is the number of categories, and $\dis{\bfx'}{\bfx}$ is the $l_0$-norm, which counts the number of different features between $\bfx$ and $\bfx'$. 
Intuitively, Eq~\ref{eq:jointdist} is the multiply of the two PMFs of the two combined approaches because the bagging and the noise addition processes are independent.

\paragraph{Distribution $\selectflip'$ for $\featfliplabelflip$.}  For $\featfliplabelflip$, we modify $\selectflip$ to $\selectflip'$ such that also flips the label of the selected instances.
Concretely, $\sample{D}$ becomes $\{(\bfx_{w_1}', y_{w_1}'), \ldots, (\bfx_{w_k}', y_{w_k}')\}$, where $y'$ is a possibly flipped label with $\symbolflip$ probability. And the PMF of $\selectflip'(D, \bfxtest)$ is,
\begin{align}
    p_{\selectflip'(D,\bfxtest)}(\sample{D}, \sample{\bfxtest}) = \frac{\symbolremain^{(k+1)d + k}}{n^k}  \left(\frac{\symbolflip}{\symbolremain}\right)^{\sum_{i=1}^k \dis{\bfx_{w_i}'}{\bfx_{w_i}} + \mathds{1}_{y_{w_i} \neq y_{w_i}'} + \dis{\bfxtest'}{\bfxtest}},
\end{align}
where $\mathds{1}_{y_{w_i} \neq y_{w_i}'}$ denotes whether the label $y_{w_i}'$ is flipped.

\paragraph{Distribution $\selectflip''$ for $\labelflip$.}  For $\labelflip$, we modify $\selectflip$ to $\selectflip''$ such that it only flips the label of the selected instances.
Also, it is not necessary to generate the smoothed test input because $\labelflip$ cannot flip input features.
Concretely, $\sample{D}$ becomes $\{(\bfx_{w_1}, y_{w_1}'), \ldots, (\bfx_{w_k}, y_{w_k}')\}$, where $y'$ is a possibly flipped label with $\symbolflip$ probability. And the PMF of $\selectflip''(D)$ is,
\begin{align}
    p_{\selectflip''(D)}(\sample{D}) = \frac{\symbolremain^{k}}{n^k}  \left(\frac{\symbolflip}{\symbolremain}\right)^{\sum_{i=1}^k \mathds{1}_{y_{w_i} \neq y_{w_i}'}}
\end{align}

\section{Neyman--Pearson Lemma for the Multi-class Case}
\label{sec: lemma_multi_class}
In multi-class case, instead of Eq~\ref{eq:certgoal1}, we need to certify 
\begin{align}
   \forall \poisoned{D} \in \poisonset{\perturb},\ \poisoned{\bfxtest} \in \perturb(\bfxtest, \ytest)_1.\ \underset{\outcome \sim \poisondist}{\PP}(\modelpredo{\outcome}=\firsty) > \underset{\outcome \sim \poisondist}{\PP}(\modelpredo{\outcome}=\secondy), \label{eq:certgoal_multi}
\end{align}
where $\secondy$ is the runner-up prediction and $\secondp$ is the probability of predicting $\secondy$. 
Formally,
\[\secondy = \underset{y\neq \firsty}{\mathrm{argmax}} \underset{\outcome \sim \cleandist}{\PP}(\modelpredo{\outcome}=y),\quad \secondp = \underset{\outcome \sim \cleandist}{\PP}(\modelpredo{\outcome}=\secondy)\]

We use Neyman--Pearson Lemma to check whether Eq~\ref{eq:certgoal_multi} holds by computing a lower bound $\lb$ of the LHS and an upper bound $\ub$ of the RHS by solving the following optimization problems,
\begin{align}
    \lb \triangleq& \min_{\sualg\in \mathcal{A}} \underset{\outcome \sim \poisondist}{\PP}(\umodelpredo{\outcome}=\firsty),\ \ub \triangleq \max_{\sualg\in \mathcal{A}} \underset{\outcome \sim \poisondist}{\PP}(\umodelpredo{\outcome}=\secondy),\label{eq:lbub_multi}\\ 
    s.t.&\ \underset{\outcome \sim \cleandist}{\PP}(\umodelpredo{\outcome}=\firsty)=\firstp \nonumber,\ \underset{\outcome \sim \cleandist}{\PP}(\umodelpredo{\outcome}=\secondy)=\secondp \nonumber\\
    &\ \poisoned{D} \in \poisonset{\perturb}, \poisoned{\bfxtest} \in \perturb(\bfxtest, \ytest)_1 \nonumber ,
\end{align}

\begin{theorem}[Neyman--Pearson Lemma for $\featfliplabelflip, \featflip, \labelflip$ in the Multi-class Case]
\label{thm:np_bound_multi_case}
Let $\poisoned{D}$ and $\poisoned{\bfx}$ be a maximally perturbed dataset and test input, i.e., $|\poisoned{D} \setminus D|=\poisonins$, $\dis{\poisoned{\bfxtest}}{\bfxtest} = \poisonfeat$, and $\dis{\poisoned{\bfxtest}_i}{\bfxtest_i} + \mathds{1}_{\poisoned{y}_i \neq y_i} = \poisonfeat$, for each perturbed example $(\poisoned{\bfxtest}_i, \poisoned{y}_i)$ in $\poisoned{D}$.
Let $i_\lb \triangleq \mathrm{argmin}_{{i \in [1,m]}} \sum_{j=1}^i \cleandistmass{\subspace_j} \ge \firstp$ and $i_\ub \triangleq \mathrm{argmax}_{{i \in [1,m]}} \sum_{j=i}^{m} \cleandistmass{\subspace_j} \ge \secondp$,
The algorithm $\sualg$ is the minimizer of Eq~\ref{eq:lbub_multi} and its behaviors among $\forall i \in \{1, \ldots, m\}\ldotp \forall \outcome \in \subspace_i$ are specified as 
\begin{small}
\begin{align*}
    \PP(\umodelpredo{\outcome} = \firsty) = \begin{cases}
    1,& i < i_\lb\\
    \frac{\firstp - \sum_{j=1}^{i-1} \cleandistmass{\subspace_j}}{\cleandistmass{\subspace_i}},& i = i_\lb \\
    0,& i > i_\lb
    \end{cases}
    \ldotp \PP(\umodelpredo{\outcome} = \secondy) = \begin{cases}
    0,& i < i_\ub\\
    \frac{\secondp - \sum_{j=i+1}^m \cleandistmass{\subspace_j}}{\cleandistmass{\subspace_i}},& i = i_\ub \\
    1,& i > i_\ub
    \end{cases}
\end{align*}
\end{small}%
Then, 
\begin{small}
\[\lb = \sum_{j=1}^{i_\lb - 1} \poisondistmass{\subspace_j} + \left(\firstp - \sum_{j=1}^{i_\lb - 1} \cleandistmass{\subspace_j}\right)/\eta_{i_\lb}, \ub = \sum_{j=i_\ub + 1}^{m} \poisondistmass{\subspace_j} + \left(\secondp - \sum_{j=i_\ub + 1}^{m} \cleandistmass{\subspace_j}\right)/\eta_{i_\ub}\]
\end{small}%
\end{theorem}

Theorem~\ref{thm:np_bound_multi_case} is a direct application of Neyman--Pearson Lemma. And Lemma 4 in \citet{tightflipbound} proves the maximally perturbed dataset and test input achieve the worst-case bound $\lb$ and $\ub$.

\subsection{Computing the Certified Radius of \technique for Perturbations $\featfliplabelflip$ and $\labelflip$}
\label{sec: other_perturbs}
\paragraph{Computing the Certified Radius for $\featfliplabelflip$}
In Eq~\ref{eq:Ldef3}, we need to consider the flipping of labels,
\begin{small}
\begin{align*}
    \underbrace{\left(\sum_{i=1}^k \dis{\bfx_{w_i}'}{\bfx_{w_i}} + \mathds{1}_{y_{w_i}' \neq y_{w_i}} + \dis{\bfxtest'}{\bfxtest}\right)}_{\diffu} - \underbrace{\left(\sum_{i=1}^k \dis{\bfx_{w_i}'}{\poisoned{\bfx}_{w_i}} + \mathds{1}_{y_{w_i}' \neq \poisoned{y}_{w_i}} +  \dis{\bfxtest'}{\poisoned{\bfxtest}}\right)}_{\diffv} = t
\end{align*}
\end{small}%
In Eq~\ref{eq:prd'}, the definition of $T(c,t)$ should be modified to,
\begin{align*}
    \sum_{\substack{0 \le u_1,\ldots, u_c \le d + 1\\ 0 \le u_0 \le d}} \sum_{\substack{0 \le v_1,\ldots, v_c \le d+1 \\ 0 \le v_0 \le d \\ u_0-v_0 + \ldots + u_c-v_c = t}} \prod_{i=1}^c L(u_i,v_i;\poisonfeat, d+1)b^{u_i}a^{d+1-u_i} L(u_0,v_0;\poisonfeat, d)b^{u_0}a^{d-u_0}
\end{align*}
The algorithm associated with Theorem~\ref{thm:selectflip_tct} should be modified as 
\begin{align*}
    T(0, t) =& \sum_{u=\max(0,t)}^{\min(d,t+d)} L(u, u-t;s, d) b^u a^{d-u},\ \forall -d \le t \le d,\\
    T(c, t) =& \sum_{t_1=\max(-d, t-cd-c+1)}^{\min(d, t+cd+c-1)} T(c-1,t - t_1)G(t_1),\ \forall c>0, -(c+1)d-c \le t \le (c+1)d+c,
\end{align*}
where $G(t)$ is defined as 
\[G(t) = \sum_{u=\max(0,t)}^{\min(d+1,t+d+1)} L(u, u-t;s, d+1) b^u a^{d+1-u},\ \forall -d-1 \le t \le d+1\]

\paragraph{Computing the Certified Radius for $\labelflip$}
In Eq~\ref{eq:Ldef3}, we only need to consider the flipping of labels,
\begin{align*}
    \underbrace{\left(\sum_{i=1}^k \mathds{1}_{y_{w_i}' \neq y_{w_i}}\right)}_{\diffu} - \underbrace{\left(\sum_{i=1}^k \mathds{1}_{y_{w_i}' \neq \poisoned{y}_{w_i}}\right)}_{\diffv} = t
\end{align*}
In the rest of the computation of $T(c,t)$, we set $d=1$, i.e., consider the label as a one-dimension feature. 

\subsection{Practical Perspectives}
\label{sec: appendix_practical}
\paragraph{Estimation of $\firstp$ and $\secondp$.}
For each test example $\bfxtest$, we need to estimate $\firstp$ and $\secondp$ of the smoothed algorithm $\salg$ given the benign dataset $D$.
We use Monte Carlo sampling to compute $\firstp$ and $\secondp$.
Specifically, for each test input $\bfxtest$, we train $N$ algorithms with the datasets $\sample{D}_1, \ldots, \sample{D}_N$ and evaluate these algorithms on input $\sample{\bfxtest}_i$, where $(\sample{D}_i, \sample{\bfxtest}_i)$ is sampled from distribution $\mu(D, \bfxtest)$.  
We count the predictions equal to label $y$ as $N_y = \sum_{i=1}^N \mathds{1}_{\modelpred{\sample{D}_i}{\sample{\bfxtest}_i}=y}$ and use the Clopper-Pearson interval to estimate $\firstp$ and $\secondp$.
\begin{align*}
    \firstp = \mathrm{Beta}(\frac{\alpha}{|\mathcal{C}|}; N_{y^*}, N-N_{y^*}+1) ,\quad 
    \secondp = \mathrm{Beta}(\frac{1 - \alpha}{|\mathcal{C}|}; N_{\secondy} + 1, N-N_{\secondy})
\end{align*}
where $1-\alpha$ is the confidence level, $|\mathcal{C}|$ is the number of different labels, and and $\mathrm{Beta}(\beta;\lambda,\theta)$ is the $\beta$th quantile of the Beta distribution of parameter $\lambda$ and $\theta$. 
We further tighten the estimation of $\secondp$ by $\min(\secondp, 1-\firstp)$ because $\firstp + \secondp \le 1$ by definition.

However, it is computationally expensive to retrain $N$ algorithms for each test input.  
We can reuse the trained $N$ algorithms to estimate $\firstp$ of $m$ test inputs with a simultaneous confidence level at least $1-\alpha$ by using \textit{Bonferroni correction}.
Specifically, we evenly divide $\alpha$ to $\frac{\alpha}{m}$ when estimating for each test input.
In the evaluation, we set $m$ to be the size of the test set.

\paragraph{Computation of the certified radius.} Previous sections have introduced how to check whether Eq~\ref{eq:certgoal1} holds for a specific $\poisonins$.
We can use binary search to find the certified radius given the estimated $\firstp$ and $\secondp$.
Although checking Eq~\ref{eq:certgoal1} has been reduced to polynomial time, it might be infeasible for in-time prediction in the real scenario.
We propose to memoize the certified radius by enumerating all possibilities of pairs of $\firstp$ and $\secondp$ beforehand so that checking Eq~\ref{eq:certgoal1} can be done in $O(1)$.
Notice that a pair of $\firstp$ and $\secondp$ is determined by $N_{\firsty}$ and $N_{\secondy}$. Recall that $N$ is the number of trained algorithms. $N_{\firsty}$ and $N_{\secondy}$ can have $O(N^2)$ different pairs ($O(N)$ in the binary-classification case).


\subsection{A Relaxation of the Neyman--Pearson Lemma}
We introduce a relaxation of the Neyman--Pearson lemma for the multi-class case.
\begin{theorem}[Relaxation of the Lemma]
\label{thm:np_non_tight_multi}
Define $\lb$ for the original subspaces $\subspace_1, \ldots, \subspace_m$ as in \cref{thm:np_bound_multi_case}.
Define $\lb_\delta$ for $\{\subspace_i\}_{i\in \indexset}$ as in \cref{thm:np_bound_multi_case}  
by underapproximating $\firstp$ as $\firstp - \sum_{i\notin \indexset} \cleandistmass{\subspace_i}$.
Define $\ub$ for the original subspaces $\subspace_1, \ldots, \subspace_m$ as in \cref{thm:np_bound_multi_case}.
Define $\ub'$ for underapproximated subspaces $\{\subspace_i\}_{i\in \indexset}$ as in \cref{thm:np_bound_multi_case} with $\secondp$ and let $\ub_\delta = \ub' + \sum_{i\notin \indexset} \poisondistmass{\subspace_i}$.
Then, we have 
\textbf{Soundness:} $\lb_\delta \le \lb, \ub_\delta \ge \ub$ and \textbf{$\delta$-Tightness:} Let $\delta \triangleq \sum_{i \notin \indexset} \poisondistmass{\subspace_i}$, then $\lb_\delta + \delta \ge \lb, \ub_\delta - \delta \ge \ub$.
\end{theorem}

\section{Proofs}
\label{sec: proofs}
\textbf{Proof of Theorem~\ref{thm:selectflip_main1}.}
\begin{proof}
If we know an outcome $\outcome \in \subspace_{c,t}$ has $\diffu$ distance from the clean data, then $\outcome$ has $(k+1)d - \diffu$ features unchanged and $\diffu$ features changed when sampling from $\cleandist$.
Thus, according to the definition of PMF in Eq~\ref{eq:jointdist}, $\cleandistmass{\outcome} = \frac{1}{n^k} \symbolremain^{(k+1)d-\diffu} \symbolflip^{\diffu}$ for all $\outcome \in \subspace_{c,t}$.
Similarly, $\poisondistmass{\outcome} = \frac{1}{n^k} \symbolremain^{(k+1)d-\diffv} \symbolflip^{\diffv}$ for all $\outcome \in \subspace_{c,t}$.
By the definition of likelihood ratios, we have $\forall \outcome \in \subspace_{c,t}.\ \eta(\outcome) = \eta_{c,t} = \cleandistmass{\outcome} / \poisondistmass{\outcome}= \symbolflip^{\diffu - \diffv} \symbolremain^{\diffv - \diffu} = \left(\frac{\symbolflip}{\symbolremain}\right)^t $.
\end{proof}

\textbf{Proof of Theorem~\ref{thm:selectflip_main2}.}
\begin{proof}
We first define a subset of $\subspace_{c,t}$ as $\subspace_{c,t,\diffu}$, 
\begin{align*}
    \subspace_{c,t,\diffu} = \{&(\{(\bfx_{w_i}', y_{w_i})\}_i, \bfx')\mid \\
    &\sum_{i=1}^k \mathds{1}_{\bfx_{w_i} \neq \poisoned{\bfx}_{w_i}}=c ,\\
    &\left(\sum_{i=1}^k \dis{\bfx_{w_i}'}{\bfx_{w_i}} + \dis{\bfx'}{\bfxtest}\right) - \left(\sum_{i=1}^k \dis{\bfx_{w_i}'}{\poisoned{\bfx}_{w_i}} + \dis{\bfx'}{\poisoned{\bfxtest}}\right) = t, \\
    & \left(\sum_{i=1}^k \dis{\bfx_{w_i}'}{\bfx_{w_i}} + \dis{\bfx'}{\bfxtest}\right) = \diffu\}
\end{align*}
Denote the size of $\subspace_{c,t,\diffu}$ as $|\subspace_{c,t,\diffu}|$, then $\cleandistmass{\subspace_{c, t}}$ can be computed as 
\begin{align}
   \cleandistmass{\subspace_{c, t}} = \sum_{0 \le \diffu \le (c+1)d} \cleandistmass{\subspace_{c, t, \diffu}} = \sum_{0 \le \diffu \le (c+1)d} \frac{1}{n^k}\symbolflip^{\diffu}\symbolremain^{d-\diffu} |\subspace_{c,t,\diffu}| \label{eq:prd_def_1}
\end{align}
Because every outcome in $\subspace_{c,t,\diffu}$ has the same probability mass and we only need to count the size of $\subspace_{c,t,\diffu}$.

\begin{align}
    |\subspace_{c,t,\diffu}| =& \binom{k}{c} \poisonins^c (n-\poisonins)^{k-c} \sum_{\substack{0 \le \diffv_0,\ldots, \diffv_c \le d\\ 0 \le \diffu_0,\ldots, \diffu_c \le d\\  \diffu_0-\diffv_0 + \ldots + \diffu_c-\diffv_c = t \\ \diffu_0 + \ldots + \diffu_c = \diffu}} \prod_{i=0}^c L(\diffu_i,\diffv_i;\poisonfeat, d),\label{eq:prd_def_2}
\end{align}
where $L(\diffu,\diffv;\poisonfeat, d)$ is defined as, similarly in the Lemma 5 of \citet{tightflipbound},
\[\sum_{i=\max(0,\diffv-\poisonfeat)}^{\min(\diffu, d-\poisonfeat, \lfloor \frac{\diffu+\diffv-\poisonfeat}{2} \rfloor)} (K-1)^j \binom{\poisonfeat}{j} \binom{\poisonfeat-j}{\diffu-i-j} K^i \binom{d-\poisonfeat}{i},\]
where $j=\diffu+\diffv-2i-\poisonfeat$.

The binomial term in Eq~\ref{eq:prd_def_2} represents the different choices of selecting the $k$ indices with $c$ perturbed indices from a pool containing $\poisonins$ perturbed indices and $n - \poisonins$ clean indices. 
The rest of Eq~\ref{eq:prd_def_2} counts the number of different choices of flips. 
Specifically, we enumerate $\diffu$ and $\diffv$ as $\sum_{i=0}^c \diffu_i$ and $\sum_{i=0}^c \diffv_i$, where $i=0$ denotes the test input. And $L(\diffu_i,\diffv_i;\poisonfeat, d)$ counts the number of different choices of flips for each example (see Lemma 5 of \citet{tightflipbound} for the derivation of $L(\diffu_i,\diffv_i;\poisonfeat, d)$).

Then, plug Eq~\ref{eq:prd_def_2} into Eq~\ref{eq:prd_def_1}, we have 
\begin{align*}
    \cleandistmass{\subspace_{c, t}} =& \sum_{0 \le \diffu \le (c+1)d} \frac{1}{n^k}\symbolflip^{\diffu}\symbolremain^{d-\diffu} \binom{k}{c} \poisonins^c (n-\poisonins)^{k-c} \sum_{\substack{0 \le \diffv_0,\ldots, \diffv_c \le d\\ 0 \le \diffu_0,\ldots, \diffu_c \le d\\  \diffu_0-\diffv_0 + \ldots + \diffu_c-\diffv_c = t \\ \diffu_0 + \ldots + \diffu_c = \diffu}} \prod_{i=0}^c L(\diffu_i,\diffv_i;\poisonfeat, d) \\
    =& \sum_{0 \le \diffu \le (c+1)d} \frac{1}{n^k} \binom{k}{c} \poisonins^c (n-\poisonins)^{k-c} \sum_{\substack{0 \le \diffv_0,\ldots, \diffv_c \le d\\ 0 \le \diffu_0,\ldots, \diffu_c \le d\\  \diffu_0-\diffv_0 + \ldots + \diffu_c-\diffv_c = t \\ \diffu_0 + \ldots + \diffu_c = \diffu}} \prod_{i=0}^c L(\diffu_i,\diffv_i;\poisonfeat, d) \symbolflip^{\diffu_i}\symbolremain^{d-\diffu_i} \\
     =& \frac{1}{n^k} \binom{k}{c} \poisonins^c (n-\poisonins)^{k-c} \sum_{\substack{0 \le \diffv_0,\ldots, \diffv_c \le d\\ 0 \le \diffu_0,\ldots, \diffu_c \le d\\  \diffu_0-\diffv_0 + \ldots + \diffu_c-\diffv_c = t}} \prod_{i=0}^c L(\diffu_i,\diffv_i;\poisonfeat, d) \symbolflip^{\diffu_i}\symbolremain^{d-\diffu_i}\\
     =&  \mathrm{Binom}(c;k,\frac{\poisonins}{n}) \sum_{\substack{0 \le \diffv_0,\ldots, \diffv_c \le d\\ 0 \le \diffu_0,\ldots, \diffu_c \le d\\  \diffu_0-\diffv_0 + \ldots + \diffu_c-\diffv_c = t}} \prod_{i=0}^c L(\diffu_i,\diffv_i;\poisonfeat, d) \symbolflip^{\diffu_i}\symbolremain^{d-\diffu_i}
\end{align*}
\end{proof}

\textbf{Proof of Theorem~\ref{thm:selectflip_tct}.}

\begin{proof}
From Eq~\ref{eq:prd'} in Theorem~\ref{thm:selectflip_main2}, then for $c\ge 1$, we have 
\begin{align*}
    T(c,t) =& \sum_{0 \le \diffu_0,\ldots, \diffu_c \le d} \sum_{\substack{0 \le \diffv_0,\ldots, \diffv_c \le d\\ \diffu_0-\diffv_0 + \ldots + \diffu_c-\diffv_c = t}} \prod_{i=0}^c L(\diffu_i,\diffv_i;\poisonfeat, d)\symbolflip^{\diffu_i}\symbolremain^{d-\diffu_i} \\
    =& \sum_{-d \le t_1 \le d} \sum_{0 \le \diffu_c \le d} \sum_{\substack{0 \le \diffv_c \le d\\ \diffu_c-\diffv_c=t_1}} L(\diffu_c,\diffv_c;\poisonfeat, d)\symbolflip^{\diffu_c}\symbolremain^{d-\diffu_c} \times \\
    & \sum_{0 \le \diffu_0,\ldots, \diffu_{c-1} \le d} \sum_{\substack{0 \le \diffv_0,\ldots, \diffv_{c-1} \le d\\ \diffu_0-\diffv_0 + \ldots + \diffu_{c-1}-\diffv_{c-1} = t-t_1}} \prod_{i=0}^{c-1} L(\diffu_i,\diffv_i;\poisonfeat, d)\symbolflip^{\diffu_i}\symbolremain^{d-\diffu_i}\\
    =& \sum_{-d \le t_1 \le d} T(0,t_1) T(c-1,t-t_1)
\end{align*}

\end{proof}

\textbf{Proof of Theorem~\ref{thm:np_non_tight}.}

Before giving the formal proof, we motivate the proof by the following knapsack problem, where each item can be divided arbitrarily. This allows a greedy algorithm to solve the problem, the same as the greedy process in Theorem~\ref{thm:np_bound}.

\begin{example}
Suppose we have $m$ items with volume $\cleandistmass{\subspace_i}$ and cost $\poisondistmass{\subspace_i}$. We have a knapsack with volume $\firstp$. Determine the best strategy to fill the knapsack with the minimal cost $\lb$. Note that each item can be divided arbitrarily.

The greedy algorithm sorts the item descendingly by ``volume per cost'' $\cleandistmass{\subspace_i}/\poisondistmass{\subspace_i}$ (likelihood ratio) and select items until the knapsack is full. The last selected item $\subspace_{i_\lb}$ will be divided to fill the knapsack. Define the best solution as $S$ in this case and the minimal cost as $\lb$.
\end{example}

Now consider Theorem~\ref{thm:np_non_tight}, which removes items $\{\subspace_i\}_{i\not\in \indexset}$ and reduces the volume of knapsack by the sum of the removed items' volume. 
Applying the greedy algorithm again, denote the best solution as $S'$ and the minimal cost in this case as $\lb_\delta$.

Soundness: The above process of removing items and reducing volume of knapsack is equivalent to just setting the cost of items in $\{\subspace_i\}_{i\not\in \indexset}$ as zero. 
Then, the new cost $\lb_\delta$ will be better than before (less than $\lb$) because the cost of some items has been set to zero.

$\delta$-tightness: If we put removed items back to the reduced knapsack solution $S'$, this new solution is a valid selection in the original problem with cost $\lb_\delta + \sum_{i\notin \indexset} \poisondistmass{\subspace_i}$, and this cost cannot be less than the minimal cost $\lb$, i.e., $\lb_\delta + \sum_{i\notin \indexset} \poisondistmass{\subspace_i} \ge \lb$.

\begin{proof}
We first consider a base case when $|B| = m-1$, i.e., only one subspace is underapproximated.
We denote the index of that subspace as $i'$.
Consider $i_\lb$ computed in Theorem~\ref{thm:np_bound}. 
\begin{itemize}
    \item If $i_\lb > i'$, then the likelihood ratio from $\{\subspace_i\}_{i\not\in \indexset}$ is used for computing $\lb$ in Theorem~\ref{thm:np_bound}, meaning $\lb_\delta = \lb - \poisondistmass{\subspace_{i'}}$. This implies both soundness and $\delta$-tightness.
    \item If $i_\lb = i'$, then $\subspace_{i'}$ partially contributes to the computation of $\lb$ in Theorem~\ref{thm:np_bound}.
    First, $\lb_\delta \le \lb$ because the computation of $\lb_\delta$ can only sum up to $i_\lb - 1$ items (it cannot sum $i_\lb$th item), which implies $\lb \ge \sum_{i=1}^{i_\lb - 1} \poisondistmass{\subspace_i} \ge \lb_\delta$.
    
    Next, we are going to prove $\lb_\delta + \poisondistmass{\subspace_{i'}} \ge \lb$.
    Suppose the additional budget $\cleandistmass{\subspace_{i'}}$ for $\lb$ selects additional subspaces (than underapproximated $\lb_\delta$) with likelihood ratio $\frac{\cleandistmass{\subspace_{i'-q}}}{\poisondistmass{\subspace_{i'-q}}}, \ldots, \frac{\cleandistmass{\subspace_{i'-1}}}{\poisondistmass{\subspace_{i'-1}}}, \frac{\cleandistmass{l_{i'}}}{\poisondistmass{l_{i'}}}$ such that $\cleandistmass{\subspace_{i'}} = \sum_{j=1}^q \cleandistmass{\subspace_{i'-j}} + \cleandistmass{l_{i'}}$, $\lb = \lb_\delta + \sum_{j=1}^q \poisondistmass{\subspace_{i'-j}} + \poisondistmass{l_{i'}}$, and $l_{i'} \subseteq \subspace_{i'}$.
    Then we have $\lb \le \lb_\delta + \poisondistmass{\subspace_{i'}}$ because $\frac{\cleandistmass{\subspace_{i'-q}}}{\poisondistmass{\subspace_{i'-q}}}\ge \ldots \ge \frac{\cleandistmass{\subspace_{i'-1}}}{\poisondistmass{\subspace_{i'-1}}} \ge \frac{\cleandistmass{l_{i'}}}{\poisondistmass{l_{i'}}} = \frac{\cleandistmass{\subspace_{i'}}}{\poisondistmass{\subspace_{i'}}}$ implies $\sum_{j=1}^q \poisondistmass{\subspace_{i'-j}} +  \poisondistmass{l_{i'}} \le \poisondistmass{\subspace_{i'}}$.
    To see this implication, we have $\poisondistmass{\subspace_{i'}} = \cleandistmass{\subspace_{i'}} / \eta_{i'} = \sum_{j=1}^q \cleandistmass{\subspace_{i'-j}} / \eta_{i'} + \cleandistmass{l_{i'}} / \eta_{i'} \ge \sum_{j=1}^q \cleandistmass{\subspace_{i'-j}} / \eta_{i'-j} + \cleandistmass{l_{i'}} / \eta_{i'} = \sum_{j=1}^q \poisondistmass{\subspace_{i'-j}} +  \poisondistmass{l_{i'}}$.
    
    \item If $i_\lb < i'$, then $\lb_\delta \le \lb$ because $\lb$ has more budget as $\firstp$ than $\firstp - \cleandistmass{\subspace_{i'}}$.
    Suppose the additional budget $\cleandistmass{\subspace_{i'}}$ for $\lb$ selects additional subspaces (than underapproximated $\lb_\delta$) with likelihood ratio $\frac{\cleandistmass{\subspace_{i_\lb - q}}}{\poisondistmass{\subspace_{i_\lb - q}}}, \ldots, \frac{\cleandistmass{\subspace_{i_\lb}}}{\poisondistmass{\subspace_{i_\lb}}}$ (we assume it selects the whole $\subspace_{i_\lb}$ for simplicity, and if it selects a subset of $\subspace_{i_\lb}$ can be proved similarly) such that $\cleandistmass{\subspace_{i'}} = \sum_{j=0}^q \cleandistmass{\subspace_{i_\lb- j}}$ and $\lb = \lb_\delta + \sum_{j=0}^q \poisondistmass{\subspace_{i_\lb - j}}$. Then we have $\lb \le \lb_\delta + \poisondistmass{\subspace_{i'}}$ because $\frac{\cleandistmass{\subspace_{i_\lb-j}}}{\poisondistmass{\subspace_{i_\lb-j}}}\ge \ldots \ge \frac{\cleandistmass{\subspace_{i_\lb}}}{\poisondistmass{\subspace_{i_\lb}}} \ge \frac{\cleandistmass{\subspace_{i'}}}{\poisondistmass{\subspace_{i'}}}$ implies $\sum_{j=0}^q \poisondistmass{\subspace_{i_\lb-j}} \le \poisondistmass{\subspace_{i'}}$. The reason of implication can be proved in a similar way as above.
\end{itemize}

We then consider $|B|<m-1$, i.e., more than one subspaces are underapproximated. 
We separate $\{\subspace_i\}_{i\not\in \indexset}$ into two parts, one $\{\subspace_{i'}\}$ contains any one of the subspaces, the other contains the rest $\{\subspace_i\}_{i\not\in \indexset \wedge i \neq i'}$. 

Denote $\lb_\delta'$ for $\{\subspace_i\}_{i\in \indexset \vee i = i'}$ as in Theorem~\ref{thm:np_bound} by underapproximating $\firstp$ as $\firstp - \sum_{i\not\in \indexset}\cleandistmass{\subspace_i} + \cleandistmass{\subspace_{i'}}$.
By inductive hypothesis, $\lb \ge \lb_\delta'$ and $\lb \le \lb_\delta' + \sum_{i\not\in \indexset}\cleandistmass{\subspace_i} - \cleandistmass{\subspace_{i'}}$.
By the same process of the above proof when one subspace is underapproximated (comparing $\{\subspace_i\}_{i\in \indexset \vee i = i'}$ with $\{\subspace_i\}_{i\in \indexset}$), we have $\lb_\delta' \ge \lb_\delta$ and $\lb_\delta' \le \lb_\delta + \cleandistmass{\subspace_{i'}}$.
Combining the results above, we have $\lb \ge \lb_\delta$ and $\lb \le \lb_\delta + \sum_{i\not\in \indexset}\cleandistmass{\subspace_i}$.

\end{proof}

\section{A KL-divergence Bound on the Certified Radius}
\label{sec: kl_bound}
We can use KL divergence~\citep{divergences} to get a looser but computationally-cheaper bound on the certified radius. Here, we certify the trigger-less case for $\featflip$.

\begin{theorem}
\label{thm:KL_bound}
Consider the binary classification case, Eq~\ref{eq:certgoal1} holds if 
\begin{align}
    \poisonins < \dfrac{n\log(4\firstp(1-\firstp))}{2k\log(\frac{\symbolflip}{\symbolremain})(\symbolremain-\symbolflip)\poisonfeat},
\end{align}
where $n$ is the size of the training set, $\poisonins$ is the certified radius, $\poisonfeat$ is the number of the perturbed features, $k$ is the size of each bag, and $\symbolremain, \symbolflip$ are the probabilities of a featuring remaining the same and being flipped. 
\end{theorem}

\begin{lemma}
\label{lemma:T}
Define $T$ as
\begin{align}
    T = \sum_{u=0}^d \sum_{v=0}^d L(u,v;\poisonfeat, d) \symbolflip^u \symbolremain^{d-u} (v-u)\label{eq:defofT}
\end{align}
where $L(u,v;\poisonfeat, d)$ is the same quantity defined in \citet{tightflipbound}. Then, we have
\[T = (\symbolremain-\symbolflip)s\]
\end{lemma}

\textbf{Proof of the Theorem~\ref{thm:KL_bound}.}
\begin{proof}
Denote $D'\sim \mu(D)$ and $D''\sim \mu(\poisoned{D})$, from the theorem in Example 5 of \cite{labelflip}, Eq~\ref{eq:certgoal1} holds if 
\begin{align}
    \KL(D''\|D') < -\frac{1}{2}\log(4\firstp(1-\firstp)) \label{eq:KLproof0}
\end{align}

Denote the PMF of selecting an index $w$ and flip $\bfx_w$ to $\bfx_w'$ by $\selectflip(D)$ as $\cleandistmass{\bfx_w'} = \frac{\symbolremain^d}{n} \left(\frac{\symbolflip}{\symbolremain}\right)^{\dis{\bfx_w'}{\bfx_w}}$.
Similarly, the PMF of selecting an index $w$ and flip $\bfx_w$ to $\bfx_w'$ by $\selectflip(\poisoned{D})$ as $\poisondistmass{\bfx_w'} = \frac{\symbolremain^d}{n} \left(\frac{\symbolflip}{\symbolremain}\right)^{\dis{\bfx_w'}{\poisoned{\bfx}_w}}$.
We now calculate the KL divergence between the distribution generated from the perturbed dataset $\poisoned{D}$ and the distribution generated from the original dataset $D$.
\begin{align}
    &\KL(D''\|D') \nonumber\\
    = &k \KL(D''_1\|D'_1)\label{eq:KLproof1}\\
    = &k \sum_{w=1}^{|D|} \sum_{\bfx_w' \in [K]^d} \poisondistmass{\bfx_w'} \log\frac{\poisondistmass{\bfx_w'}}{\cleandistmass{\bfx_w'}}\nonumber\\
    = &k \sum_{\bfx_w \neq \poisoned{\bfx}_w} \sum_{\bfx_w' \in [K]^d} \poisondistmass{\bfx_w'} \log\frac{\poisondistmass{\bfx_w'}}{\cleandistmass{\bfx_w'}}\label{eq:KLproof2}\\
    \le &k\poisonins \sum_{\bfx_w' \in [K]^d} \frac{\symbolremain^d}{n}  \left(\frac{\symbolflip}{\symbolremain}\right)^{\dis{\bfx_w'}{\poisoned{\bfx}_w}} \log(\frac{\symbolflip}{\symbolremain}) (\dis{\bfx_w'}{\poisoned{\bfx}_w} - \dis{\bfx_w'}{\bfx_w})\nonumber\\
    = & k\frac{\poisonins}{n} \symbolremain^d \log(\frac{\symbolflip}{\symbolremain}) \sum_{\bfx_w' \in [K]^d}  \left(\frac{\symbolflip}{\symbolremain}\right)^{\dis{\bfx_w'}{\poisoned{\bfx}_w}} (\dis{\bfx_w'}{\poisoned{\bfx}_w} - \dis{\bfx_w'}{\bfx_w})\label{eq:KLproof3}
\end{align}
where $\KL(D_1'' \| D_1')$ is the KL divergence of the first selected instance.
We have Eq~\ref{eq:KLproof1} because each selected instance is independent. 
We have Eq~\ref{eq:KLproof2} because the $\cleandistmass{\bfx_w'}$ and $\poisondistmass{\bfx_w'}$ only differs when the $w$th instance is perturbed.

The attacker can modify $\poisoned{D}$ to maximize Eq~\ref{eq:KLproof3}. And the Lemma 4 in \citet{tightflipbound} states that the maximal value is achieved when $\poisoned{\bfx}_w$ has exact $\poisonfeat$ features flipped to another value. 
Now suppose there are $\poisonfeat$ features flipped in $\poisoned{\bfx}_w$, we then need to compute Eq~\ref{eq:KLproof3}. 
If we denote $\dis{\bfx_w'}{\poisoned{\bfx}_w}$ as $u$ and $\dis{\bfx_w'}{\bfx_w}$ as $v$, and we count the size of the set $L(u,v;\poisonfeat,d )=\{\bfx_w' \in [K]^d \mid \dis{\bfx_w'}{\poisoned{\bfx}_w} = u, \dis{\bfx_w'}{\bfx_w} = v, \dis{\bfx_w}{\poisoned{\bfx}_w} = \poisonfeat\}$, then we can compute Eq~\ref{eq:KLproof3} as
\begin{align*}
   k\frac{\poisonins}{n} \symbolremain^d \log(\frac{\symbolremain}{\symbolflip}) \sum_{u=0}^d \sum_{v=0}^d |L(u,v;\poisonfeat, d)| \left(\frac{\symbolflip}{\symbolremain}\right)^{u}(v-u) 
\end{align*}
Let $T$ be defined as in Eq~\ref{eq:defofT}, we then have
\begin{align}
    \KL(D'' \| D') = k\frac{\poisonins}{n} \log(\frac{\symbolremain}{\symbolflip})T \label{eq:KLproof4}
\end{align}
Combine Eq~\ref{eq:KLproof0}, Eq~\ref{eq:KLproof4}, and Lemma~\ref{lemma:T}, we have 
\begin{align*}
    k\frac{\poisonins}{n} \log(\frac{\symbolremain}{\symbolflip})(\symbolremain-\symbolflip)\poisonfeat \le \KL(D''\|D') <& -\frac{1}{2}\log(4\firstp(1-\firstp))\\
    \poisonins  <& \dfrac{n\log(4\firstp(1-\firstp))}{2k\log(\frac{\symbolflip}{\symbolremain})(\symbolremain-\symbolflip) \poisonfeat}
\end{align*}
\end{proof}

\textbf{Proof of the Lemma~\ref{lemma:T}.}
\begin{proof}
Notice that the value of $T$ does not defend on the feature dimension $d$. 
Thus, we prove the lemma by induction on $d$.
We further denote the value of $T$ under the feature dimension $d$ as $T_d$.

Let $L(u,v;\poisonfeat, d)$ be defined as in the Lemma 5 of \citet{tightflipbound},
\[\sum_{i=\max(0,v-\poisonfeat)}^{\min(u, d-\poisonfeat, \lfloor \frac{u+v-\poisonfeat}{2} \rfloor)} (K-1)^j \binom{\poisonfeat}{j} \binom{\poisonfeat-j}{u-i-j} K^i \binom{d-\poisonfeat}{i},\]
where $j=u+v-2i-\poisonfeat$.

\textbf{Base case.} Because $0 \le s \le d$, when $d = 0$, it is easy to see $T_0 = s(\symbolremain-\symbolflip) = 0$.

\textbf{Induction case.} 
We first prove $T_{d+1}=s(\symbolremain-\symbolflip)$ under a special case, where $s = d+1$, given the inductive hypothesis $T_d = s(\symbolremain-\symbolflip)$ for  $s = d$. 

By definition
\begin{align}
    T_{d+1} = \sum_{u=0}^{d+1} \sum_{v=0}^{d+1} |L(u,v;\poisonfeat,d+1)| \symbolflip^u \symbolremain^{d+1-u} (v-u),\label{eq:lemmaT0}
\end{align}

By the definition of $L(u,v;\poisonfeat,d)$, we have the following equation when $0 \le s \le d$,
\begin{align}
    |L(u,v;d+1,d+1)| = (K-1)|L(u-1,v-1;d,d)| +|L(u-1,v;d,d)| + |L(u,v-1;d,d)|\label{eq:lemmaT4}
\end{align}
and 
\begin{align}
    \sum_{u=0}^{d} \sum_{v=0}^{d} |L(u,v;d,d)| \symbolflip^u \symbolremain^{d-u} = 1\label{eq:lemmaT8}
\end{align}
Plug Eq~\ref{eq:lemmaT4} into Eq~\ref{eq:lemmaT0}, we have the following equations when $s = d+1$,
\begin{align}
    T_{d+1} = &\sum_{u=0}^{d+1} \sum_{v=0}^{d+1} (K-1)|L(u-1,v-1;d,d)|\symbolflip^u \symbolremain^{d+1-u} (v-u) + \nonumber\\
    &\sum_{u=0}^{d+1} \sum_{v=0}^{d+1} |L(u-1,v;d,d)| \symbolflip^u \symbolremain^{d+1-u}  (v-u)+\nonumber\\
    &\sum_{u=0}^{d+1} \sum_{v=0}^{d+1} |L(u,v-1;d,d)| \symbolflip^u \symbolremain^{d+1-u} (v-u)\nonumber\\
    = &\symbolflip(K-1)\sum_{u=0}^{d} \sum_{v=0}^{d} |L(u,v;d,d)|\symbolflip^u \symbolremain^{d-u} (v-u) + \label{eq:lemmaT5}\\
    &\symbolflip\sum_{u=0}^{d} \sum_{v=0}^{d} |L(u,v;d,d)| \symbolflip^{u} \symbolremain^{d-u} (v-u-1)+\label{eq:lemmaT6}\\
    &\symbolremain\sum_{u=0}^{d} \sum_{v=0}^{d} |L(u,v;d,d)| \symbolflip^{u} a^{d-u} (v-u+1)\label{eq:lemmaT7}\\
    = & \sum_{u=0}^{d} \sum_{v=0}^{d} |L(u,v;d,d)| \symbolflip^{u} \symbolremain^{d-u} [(v-u) + (\symbolremain-\symbolflip)] \label{eq:lemmaT10}\\
    = &d(\symbolremain-\symbolflip) + (\symbolremain-\symbolflip)\label{eq:lemmaT9}\\
    = &\poisonfeat(\symbolremain-\symbolflip)\nonumber
\end{align}
We have Eq~\ref{eq:lemmaT4}, Eq~\ref{eq:lemmaT5} and Eq~\ref{eq:lemmaT6} because $L(u,v;\poisonfeat,d)=0$ when $u>d$, $v>d$, $u<0$ , or $v<0$.
We have Eq~\ref{eq:lemmaT10} because $\symbolflip(K-1) + \symbolflip + \symbolremain = 1$ as $\symbolflip$ is defined as $\frac{1 - \symbolremain}{K}$.
We have Eq~\ref{eq:lemmaT9} by plugging in Eq~\ref{eq:lemmaT8}.

Next, we are going to show that $T_{d+1} = s(\symbolremain - \symbolflip)$ for all $0 \le s \le d$, given the inductive hypothesis $T_d = s(\symbolremain - \symbolflip)$ for all $0 \le s \le d$. 

By the definition of $L(u,v;\poisonfeat,d)$, we have the following equations when $0 \le s \le d$,
\begin{align}
    |L(u,v;\poisonfeat,d+1)| = |L(u,v;\poisonfeat,d)| + K|L(u-1,v-1;\poisonfeat,d)|\label{eq:lemmaT1}
\end{align}
Plug Eq~\ref{eq:lemmaT1} into Eq~\ref{eq:lemmaT0}, for all $0 \le s \le d$, we have
\begin{align}
    T_{d+1} = &\sum_{u=0}^{d+1} \sum_{v=0}^{d+1} |L(u,v;\poisonfeat,d)|\symbolflip^u \symbolremain^{d+1-u} (v-u) + \nonumber\\
    &\sum_{u=0}^{d+1} \sum_{v=0}^{d+1} K|L(u-1,v-1;\poisonfeat,d)| \symbolflip^u \symbolremain^{d+1-u} (v-u)\nonumber\\
    = &\symbolremain\sum_{u=0}^{d} \sum_{v=0}^{d} |L(u,v;\poisonfeat,d)|\symbolflip^u \symbolremain^{d-u} (v-u) + \label{eq:lemmaT2}\\
    &\symbolflip K\sum_{u=0}^{d} \sum_{v=0}^{d} |L(u,v;\poisonfeat,d)| \symbolflip^{u} \symbolremain^{d-u} (v-u)\label{eq:lemmaT3}\\
    = & \symbolremain\poisonfeat(\symbolremain-\symbolflip) + \symbolflip K\poisonfeat(\symbolremain-\symbolflip) \nonumber\\
    = &\poisonfeat(\symbolremain-\symbolflip)\nonumber
\end{align}
We have Eq~\ref{eq:lemmaT2} and Eq~\ref{eq:lemmaT3} because $L(u,v;\poisonfeat,d)=0$ when $u>d$, $v>d$, $u<0$ , or $v<0$.

\end{proof}

\section{Experiments}
\label{sec: appendix_exp}
\subsection{Dataset Details}
\label{sec: appendix_data}
MNIST is an image classification dataset containing 60,000 training and 10,000 test examples. Each example can be viewed as a vector containing 784~($28\times 28$) features.

CIFAR10 is an image classification dataset containing 50,000 training and 10,000 test examples. Each example can be viewed as a vector containing 3072~($32\times 32 \times 3$) features.

EMBER is a malware detection dataset containing 600,000 training and 200,000 test examples. Each example is a vector containing 2,351 features.

Contagio is a malware detection dataset, where each example is a vector containing 135 features. We partition the dataset into 6,000 training and 4,000 test examples. 

MNIST-17 is a sub-dataset of MNIST, which contains 13,007 training and 2,163 test examples. MNIST-01 is a sub-dataset of MNIST, which contains 12,665 training and 2,115 test examples. CIFAR10-02 is a sub-dataset of CIFAR10, which contains 10,000 training and 2,000 test examples. 

For CIFAR10, we categorize each feature into 4 categories. For the rest of the datasets, we binarize each feature.
A special case is $\labelflip$, where we do not categorize features.

\begin{figure}[h]
\centering
    \hspace{-2.5em}
    \subfigure[$\featflipone$ on MNIST]{
        \includegraphics[width=.55\textwidth]{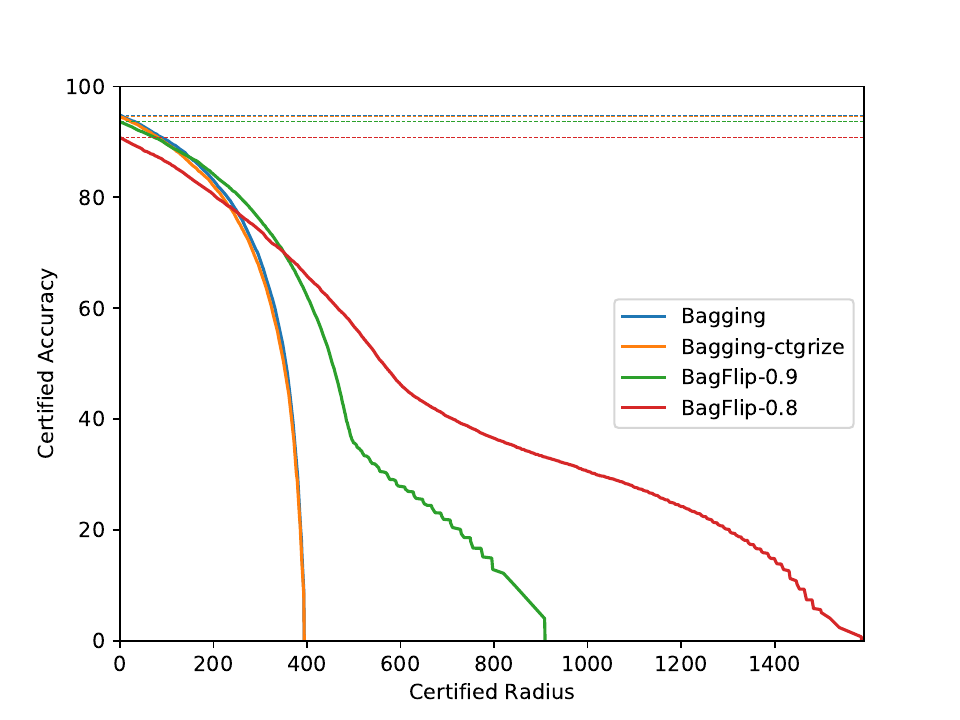}
        }%
    ~    
    \hspace{-3em}
    \subfigure[$\featflipinf$ on MNIST]{
        \includegraphics[width=.55\textwidth]{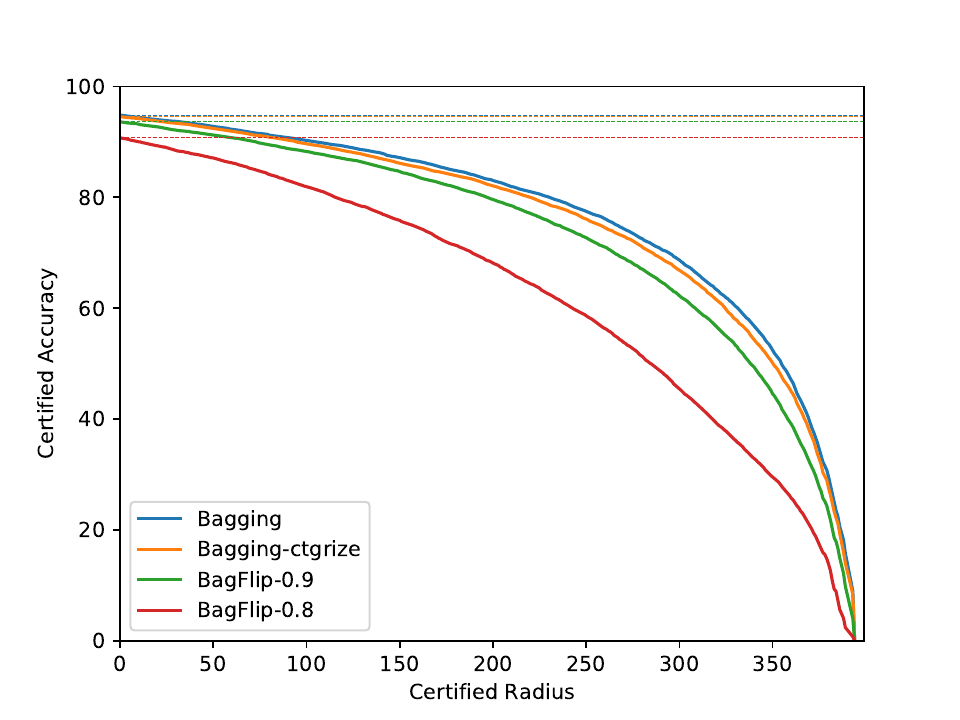}
        }
        
    \vspace{-1.1em}
    
    \hspace{-2.5em}
    \subfigure[$\featflipone$ on EMBER]{
        \includegraphics[width=.55\textwidth]{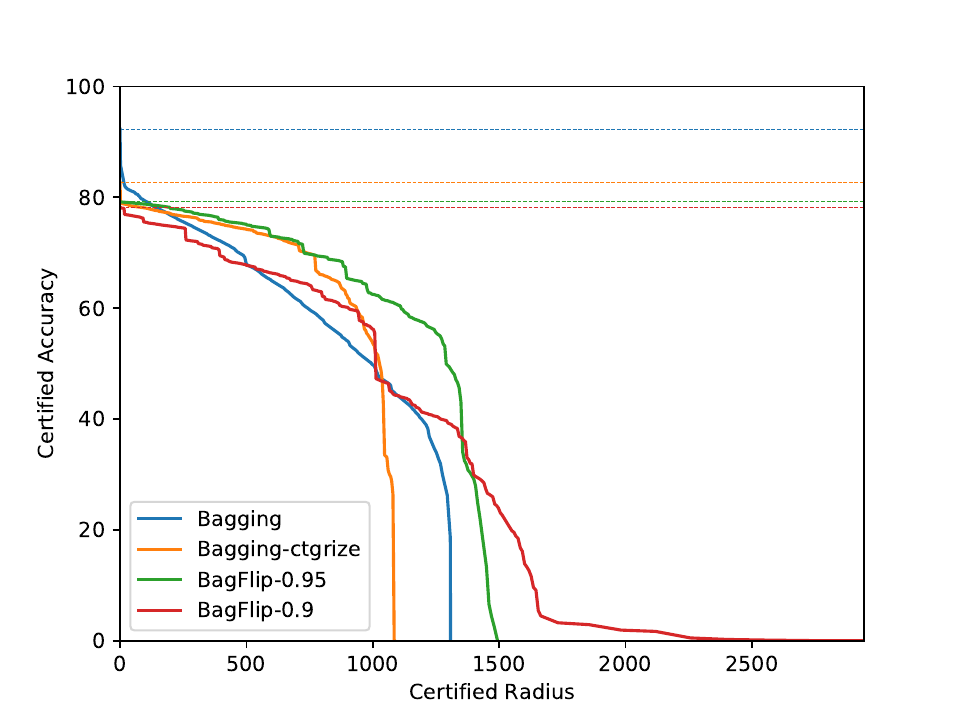}
        }%
    ~
    \hspace{-3em}
    \subfigure[$\featflipinf$ on EMBER]{
        \includegraphics[width=.55\textwidth]{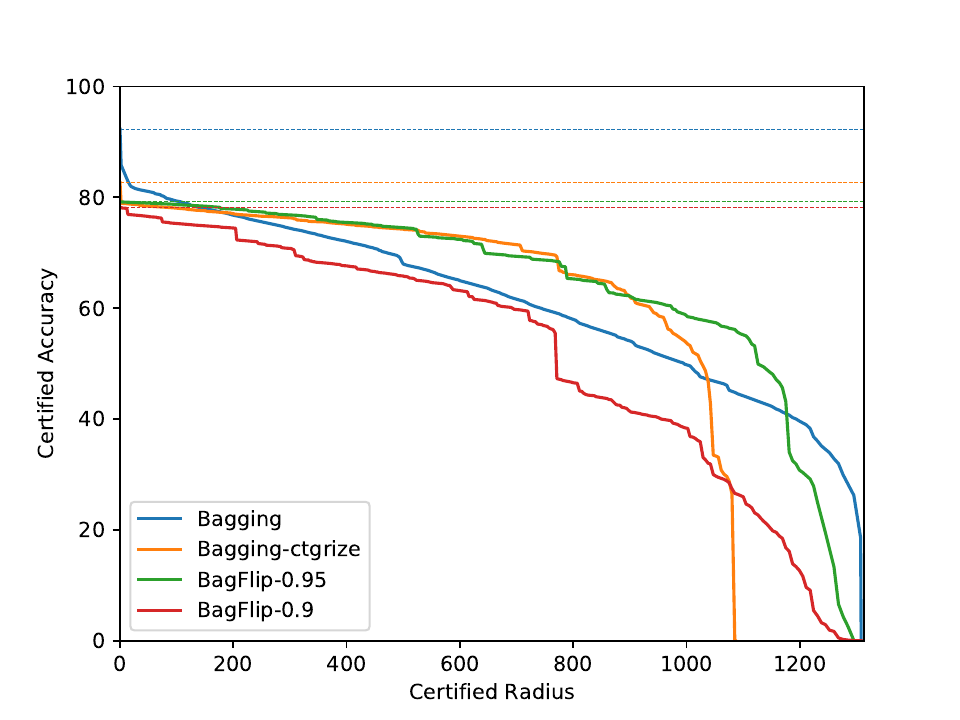}
        }
    \vspace{-1.1em}
    
    \hspace{-2.5em}
    \subfigure[$\featflipone$ on Contagio]{
        \includegraphics[width=.55\textwidth]{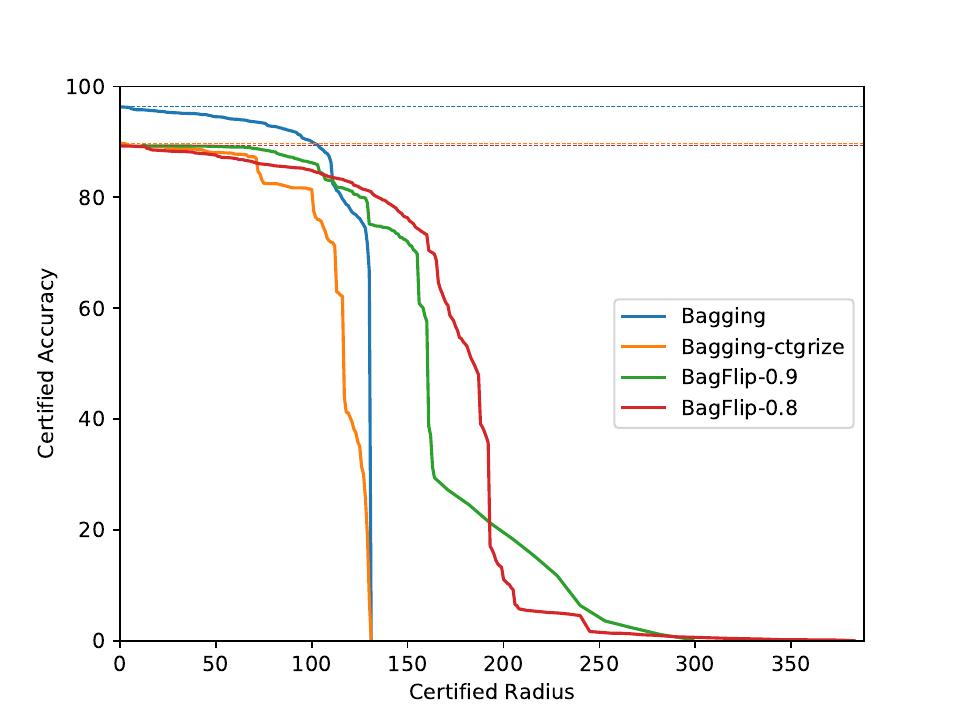}
        }%
    ~
    \hspace{-3em}
    \subfigure[$\featflipinf$ on Contagio]{
        \includegraphics[width=.55\textwidth]{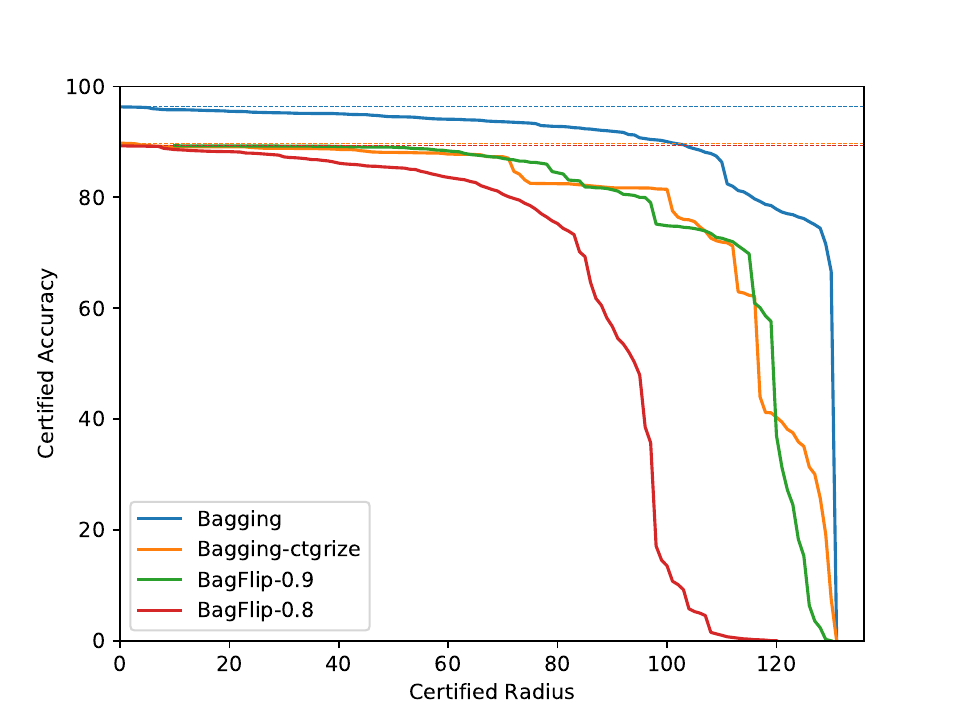}
        }
\caption{Compared to Bagging~\cite{bagging}. The horizontal dashed lines show the normal accuracy. The solid lines show the certified accuracy at different $\assumeradii$. \technique-$a$ shows the result of the noise level $a$. The blue line shows the uncategorized version of bagging.}
\label{fig:compare_bagging_full}
\end{figure}

\begin{figure}[h]
\centering
    \hspace{-2.5em}
    \subfigure[\technique-0.9 on MNIST]{
        \includegraphics[width=.55\textwidth]{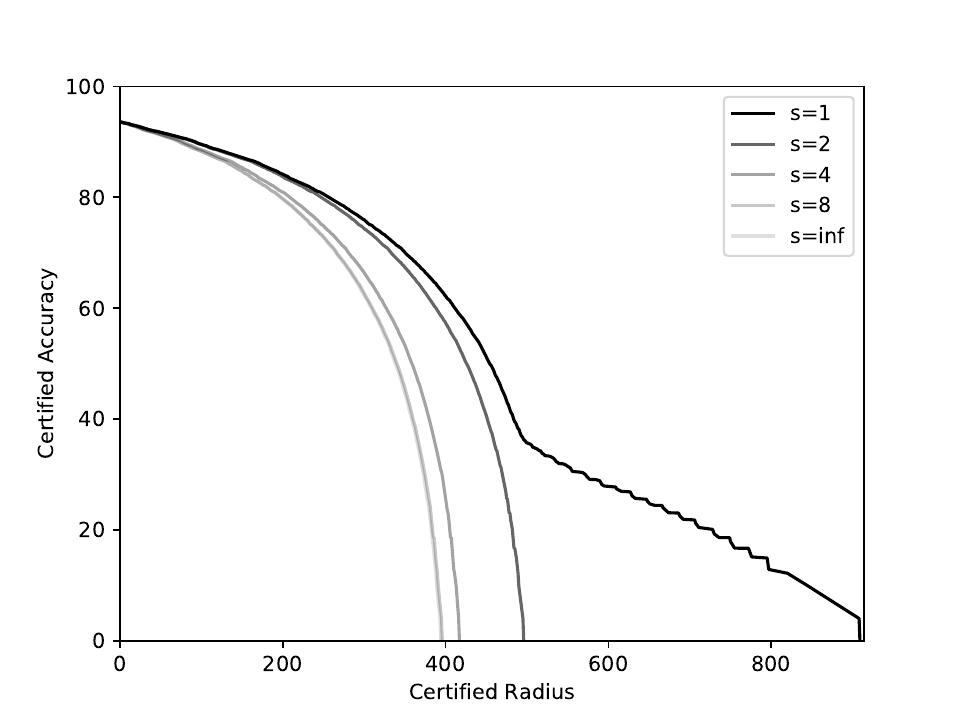}
        }%
    ~    
    \hspace{-3em}
    \subfigure[\technique-0.8 on MNIST]{
        \includegraphics[width=.55\textwidth]{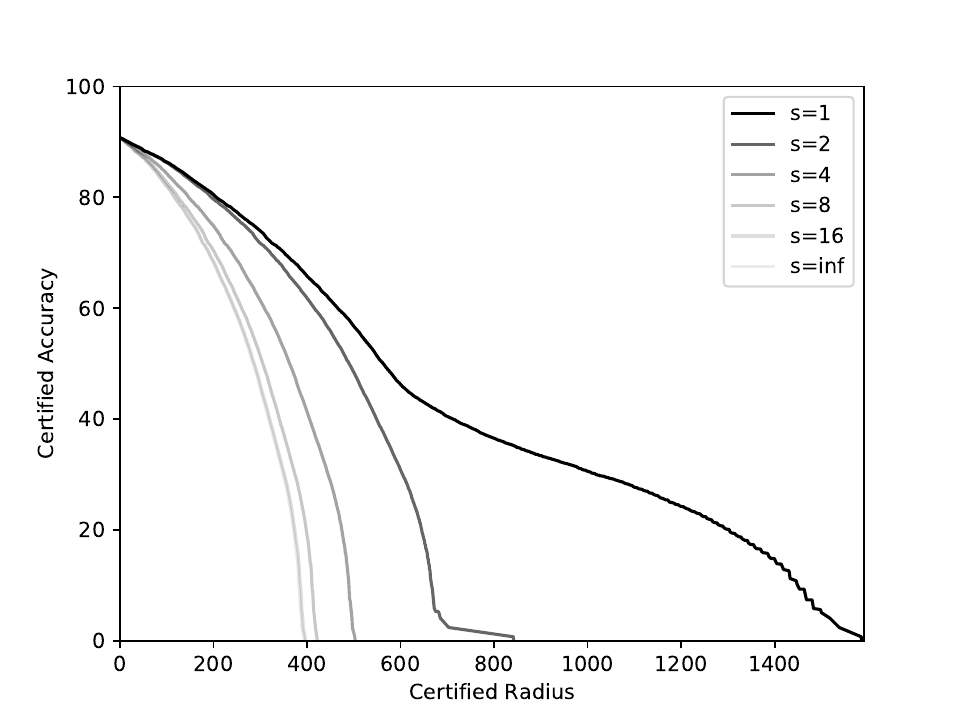}
        }
    \vspace{-1.1em}
        
    \hspace{-2.5em}
    \subfigure[\technique-0.95 on EMBER]{
        \includegraphics[width=.55\textwidth]{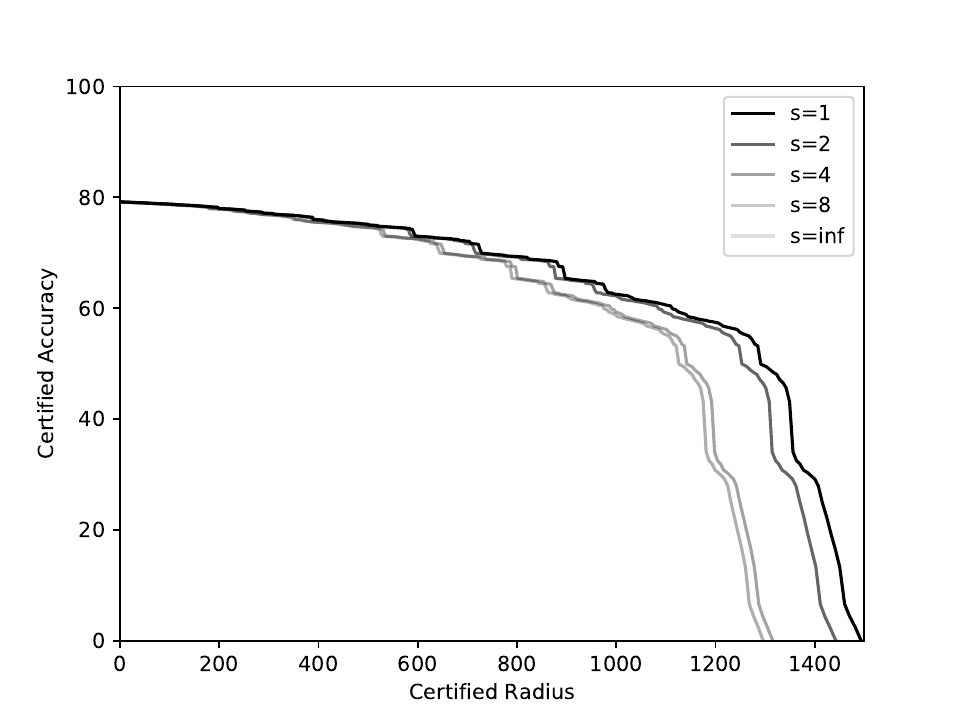}
        }%
    ~
    \hspace{-3em}
    \subfigure[\technique-0.9 on EMBER]{
        \includegraphics[width=.55\textwidth]{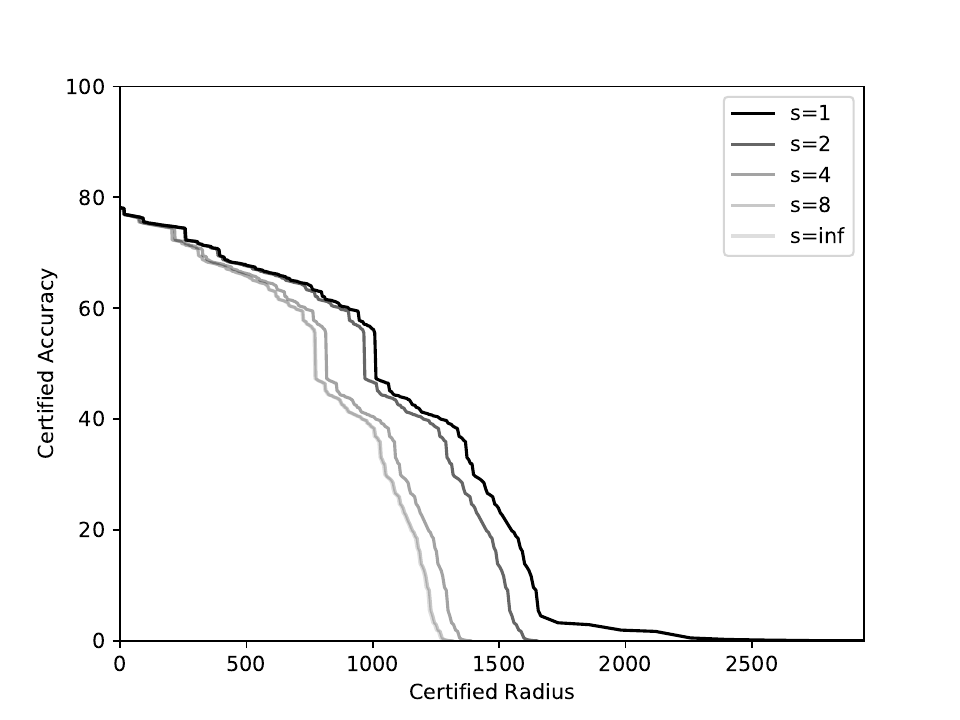}
        }
    \vspace{-1.1em}
    
    \hspace{-2.5em}
    \subfigure[\technique-0.9 on Contagio]{
        \includegraphics[width=.55\textwidth]{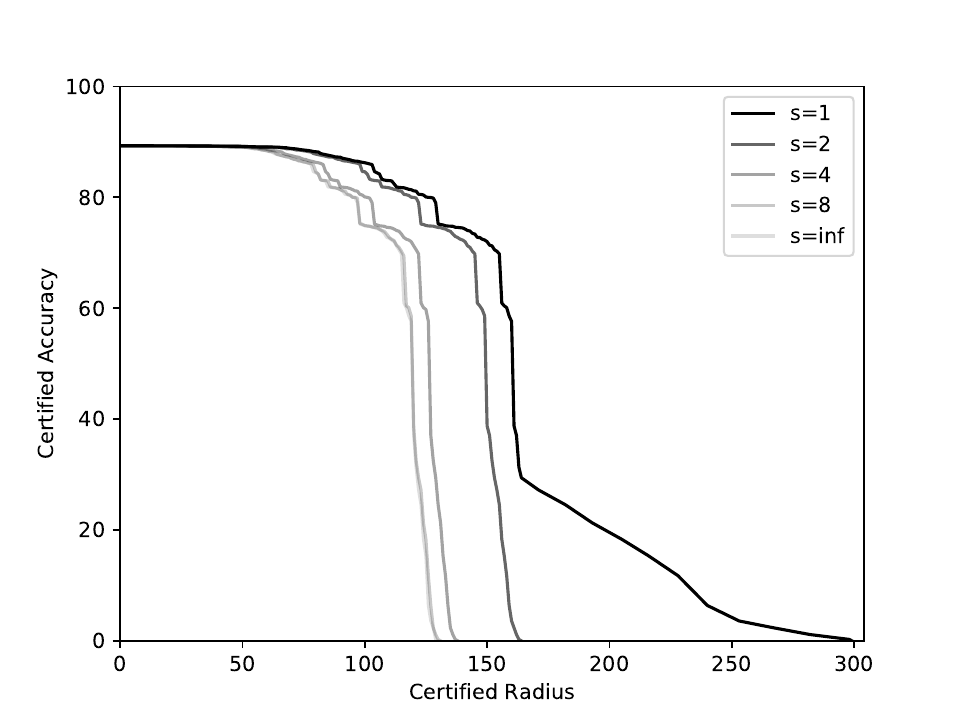}
        }%
    ~
    \hspace{-3em}
    \subfigure[\technique-0.8 on Contagio]{
        \includegraphics[width=.55\textwidth]{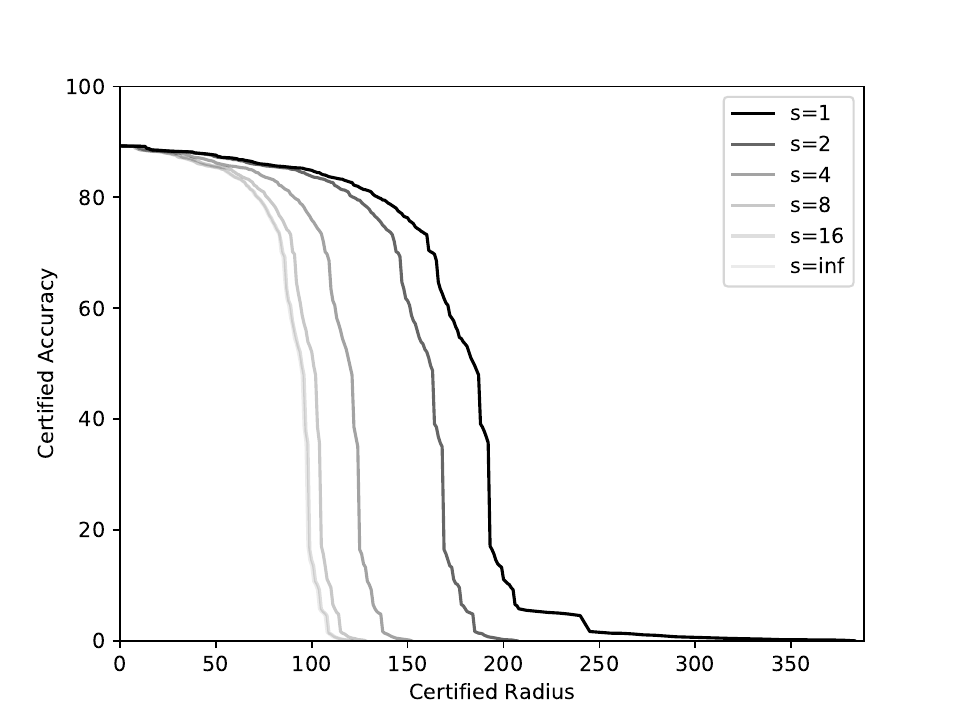}
        }
\caption{Results of \technique on different $\poisonfeat$ and different datasets.}
\label{fig:different_s_full}
\end{figure}

\begin{figure}[h]
\centering
    \hspace{-2.5em}
    \subfigure[$\featflipone$ on CIFAR10]{
        \includegraphics[width=.55\textwidth]{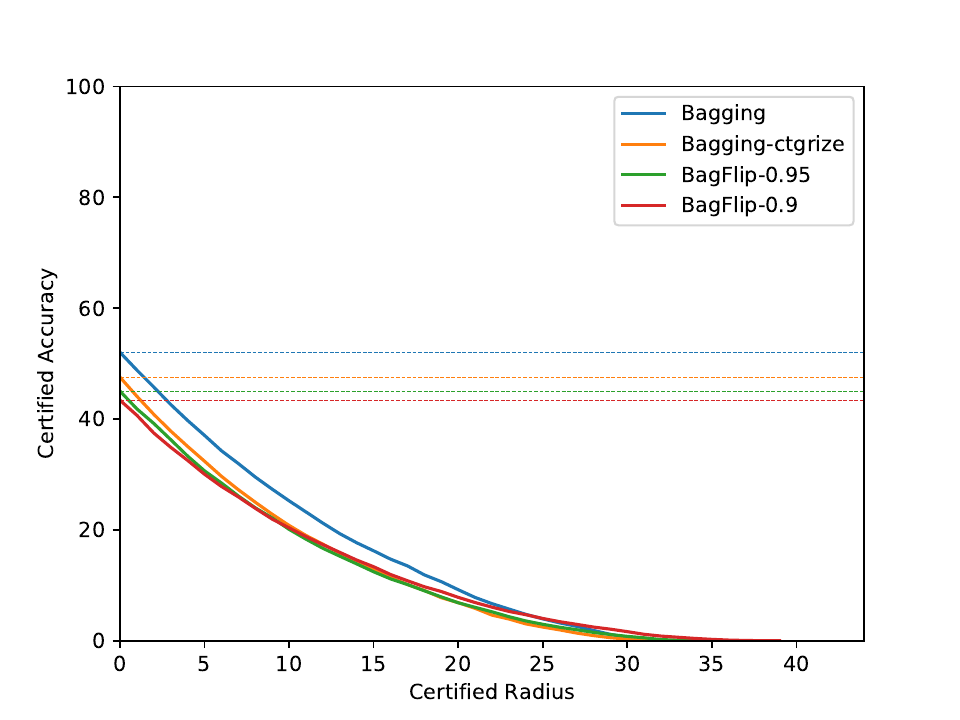}
        }%
    ~    
    \hspace{-3em}
    \subfigure[$\featflipone$ on CIFAR10]{
        \includegraphics[width=.55\textwidth]{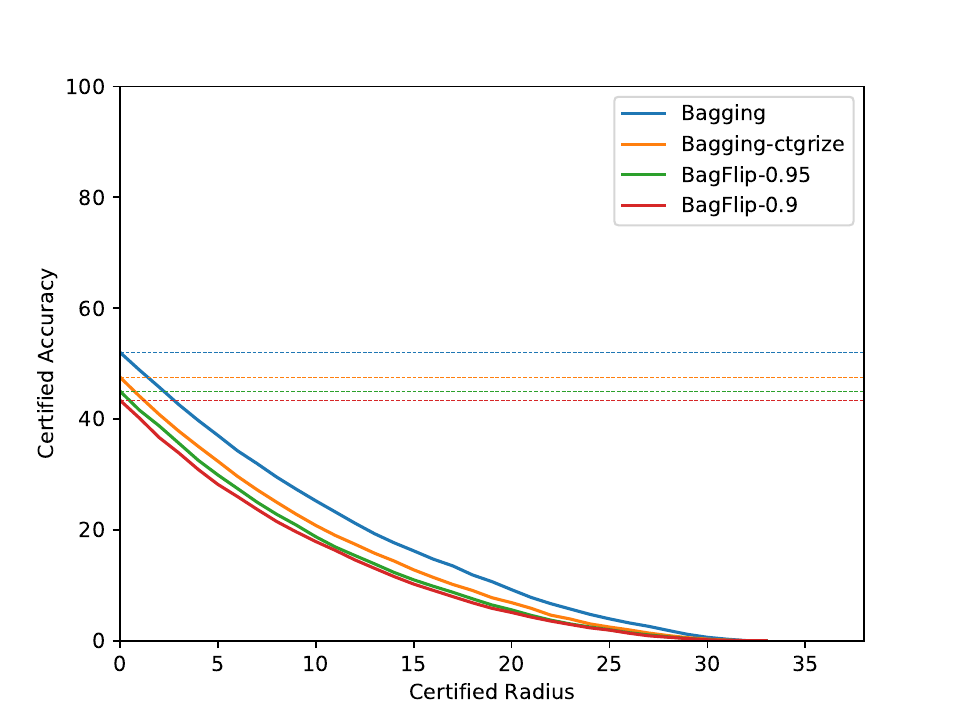}
        }
    \vspace{-1.1em}
    
    \hspace{-2.5em}
    \subfigure[\technique-0.95 on CIFAR10]{
        \includegraphics[width=.55\textwidth]{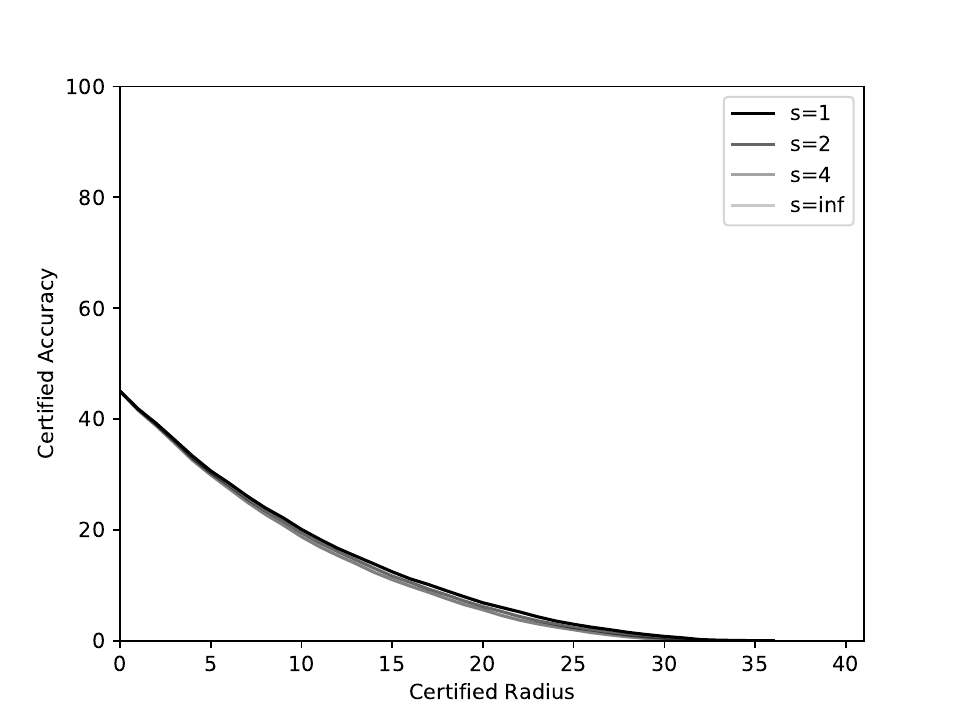}
        }%
    ~
    \hspace{-3em}
    \subfigure[\technique-0.9 on CIFAR10]{
        \includegraphics[width=.55\textwidth]{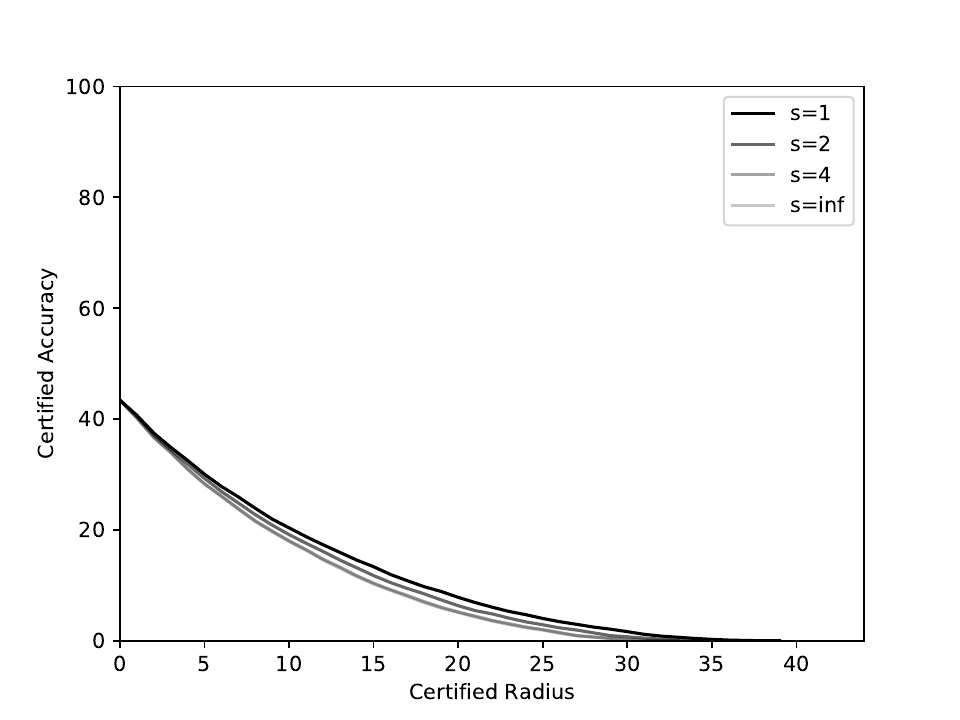}
        }
\caption{Results of \technique on CIFAR10.}
\label{fig:cifar10}
\end{figure}

\subsection{Experiment Results Details}
\label{sec: appendix_res}
\begin{table}[t]
    \centering
    \caption{This paper compared to other approaches. RAB~\cite{RAB} can handle perturbation $\featflip^*$ that perturbs the input within a $l_2$-norm ball of radius $\poisonfeat$.}
 \renewcommand{\arraystretch}{1}
    \begin{small}
    \begin{tabular}{llllc}
        \toprule
         Approach & Perturbation $\perturb$ & Probability measure $\mu$ & Goal\\ 
        \midrule
        \citet{bagging, framework_def} & Any & Bagging  & Trigger-less \\
        \citet{labelflip} & $\labelflip$ & Noise  & Trigger-less  \\
        RAB~\cite{RAB} & $\featflip^*$ & Noise  & Trigger-less, Backdoor  \\
        \citet{featureflip} & $\featfliplabelflip, \featflip, \labelflip$ & Noise  & Trigger-less, Backdoor  \\
        \emph{This paper} & $\featfliplabelflip, \featflip, \labelflip$ & Bagging+Noise & Trigger-less, Backdoor \\
        \bottomrule
    \end{tabular}
    \end{small}
    \vspace{-1em}
    \label{tab:approaches}
\end{table}

We train all models on a server running Ubuntu 18.04.5 LTS with two V100 32GB GPUs and Intel Xeon Gold 5115 CPUs running at 2.40GHz.
For computing the certified radius, we run experiments across hundreds of machines in high throughput computing center.

\subsubsection{Defend Trigger-less Attacks}

\paragraph{Comparison to Bagging} We show full results of comparison on $\featflip$ in Figures~\ref{fig:compare_bagging_full}~\ref{fig:different_s_full}~and~\ref{fig:cifar10}.
The results are similar as described in \Cref{sec: experiment}. 

\begin{figure}[h]
\centering
    \hspace{-2.5em}
    \subfigure[Compared to Bagging on $\featfliplabelflipone$ on MNIST-17]{
        \includegraphics[width=.55\textwidth]{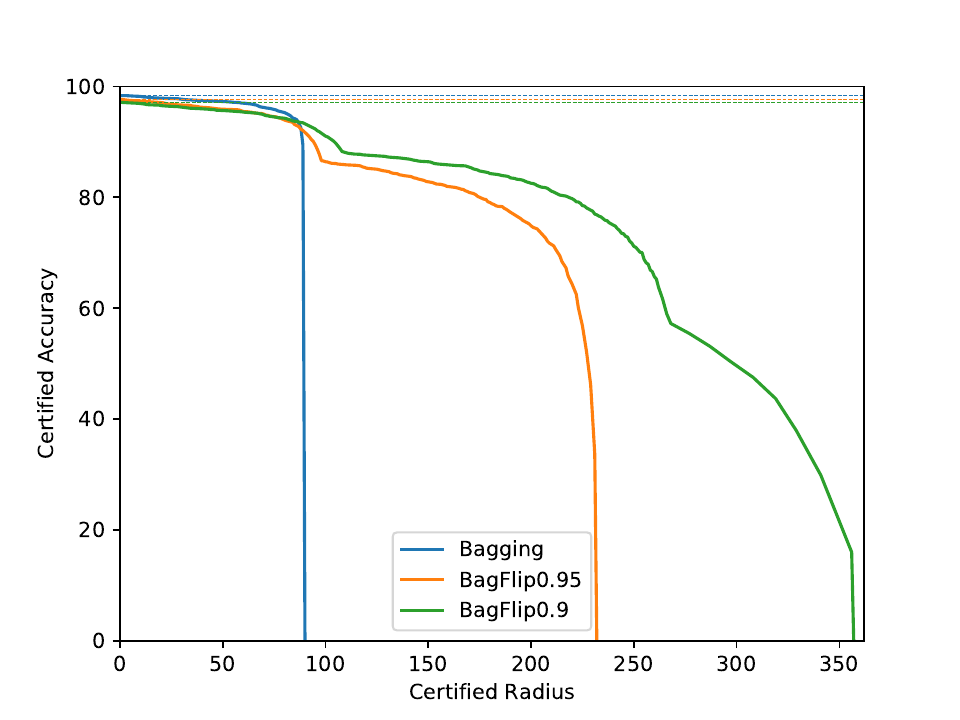}
        }%
    ~    
    \hspace{-3em}
    \subfigure[Compared to Bagging on $\labelflip$ using MNIST]{
        \includegraphics[width=.55\textwidth]{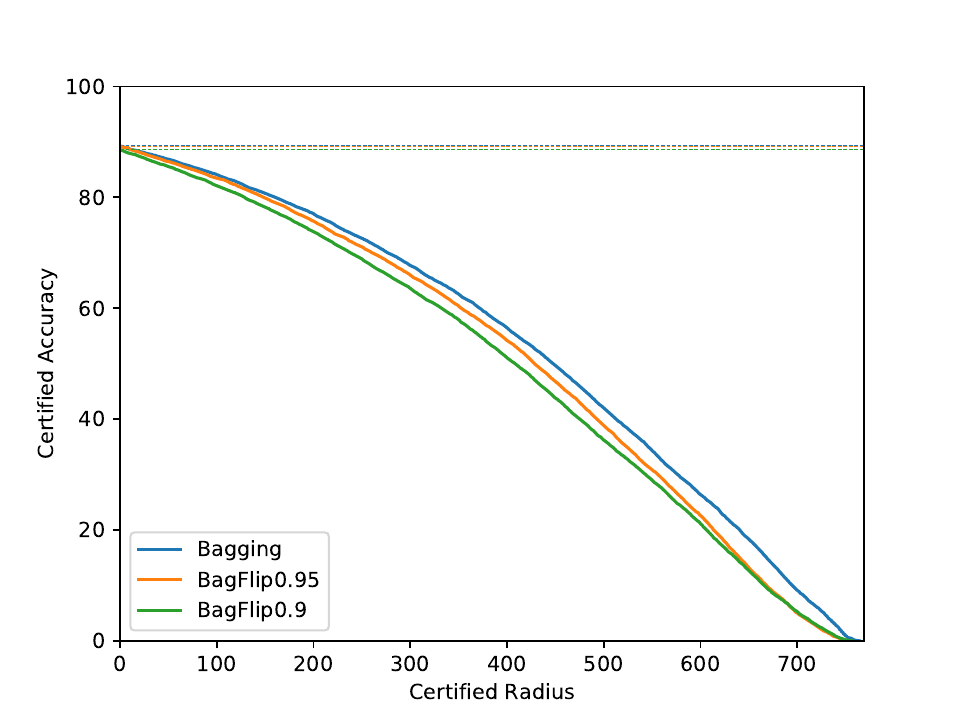}
        }
\caption{Compared to Bagging on $\featfliplabelflipone$ on MNIST-17 and $\labelflip$ using MNIST.}
\label{fig:other_perturbs_bagging}
\end{figure}

We additionally compare \technique with Bagging on $\featfliplabelflipone$ using MNIST-17 and $\labelflip$ using MNIST, and show the results in Figure~\ref{fig:other_perturbs_bagging}. \technique still outperforms Bagging on $\featfliplabelflipone$ using MNIST-17.
However, Bagging outperforms \technique on $\labelflip$ because when the attacker is only able to perturb the label, then $\poisonfeat = 1$ is equal to $\poisonfeat = \infty$ and flipping the labels hurts the accuracy. 


\begin{figure}[t]
    \centering
     \includegraphics[width=0.75\columnwidth]{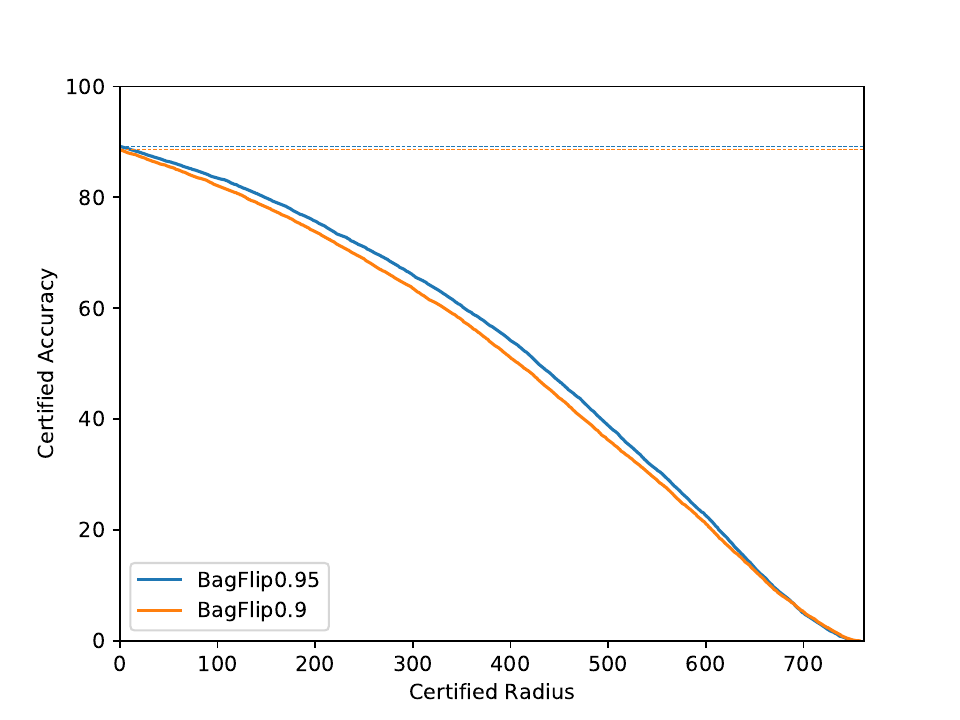}
    \caption{Results of \technique on $\labelflip$.}
    \label{fig:compare_label}
\end{figure}

\paragraph{Comparison to LabelFlip} 
We compare two configurations of \technique to LabelFlip using MNIST and show results of \technique in Figure~\ref{fig:compare_label}. 
The results show that LabelFlip achieves less than $60\%$ normal accuracy, while \technique-0.95~(\technique-0.9) achieves 89.2\%~(88.6\%) normal accuracy, respectively.
\technique achieves higher certified accuracy than LabelFlip across all $\assumeradii$.
In particular, the certified accuracy of LabelFlip drops to less than $20\%$ when $\assumeradii=0.83$, while \technique-0.95~(\technique-0.9) still achieves 38.9\%~(36.2\%) certified accuracy, respectively.
\textbf{\technique has higher normal accuracy and certified accuracy than LabelFlip.}

\subsubsection{Defend Backdoor Attacks}
We set $k=100$ for MNIST-17 when comparing to FeatFlip. We set $k=50, 200$ for MNIST-01 and CIFAR10-02 respectively when comparing to RAB.
And we set $k=100, 1000, 3000, 30$ for MNIST, CIFAR10, EMBER, and Contagio respectively when evaluating \technique on $\featflipone$.
We use BadNets to modify 10\% of examples in the training set.



\begin{figure}[h]
\centering
    \hspace{-2.5em}
    \subfigure[Compared to RAB on MNIST-01]{
        \includegraphics[width=.55\textwidth]{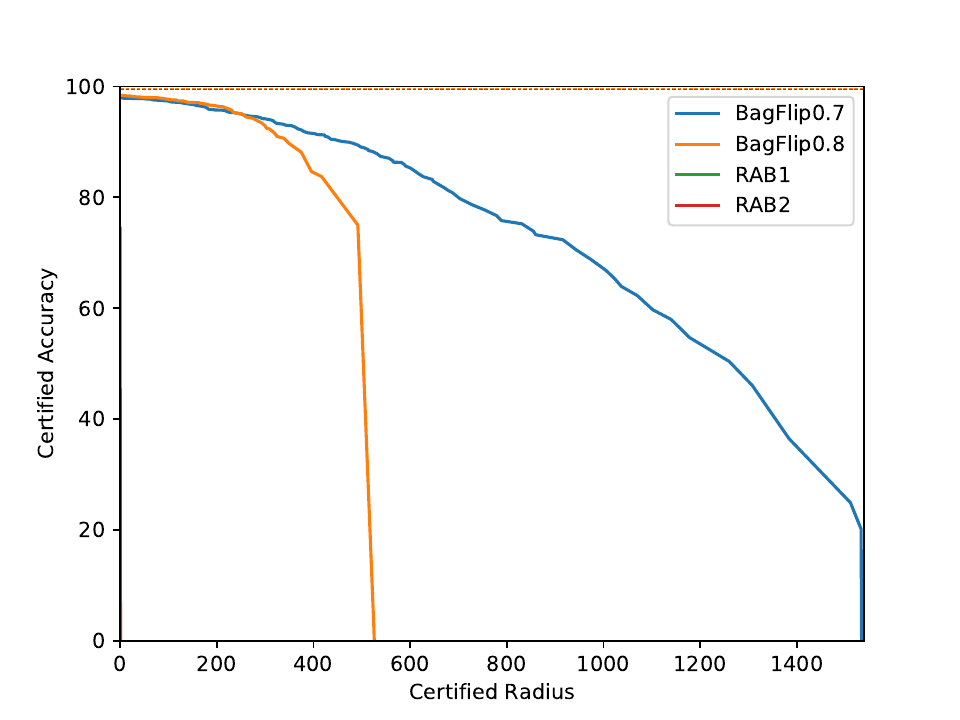}
        \label{fig:rab1}
        }%
    ~    
    \hspace{-3em}
    \subfigure[Compared to RAB on CIFAR10-02]{
        \includegraphics[width=.55\textwidth]{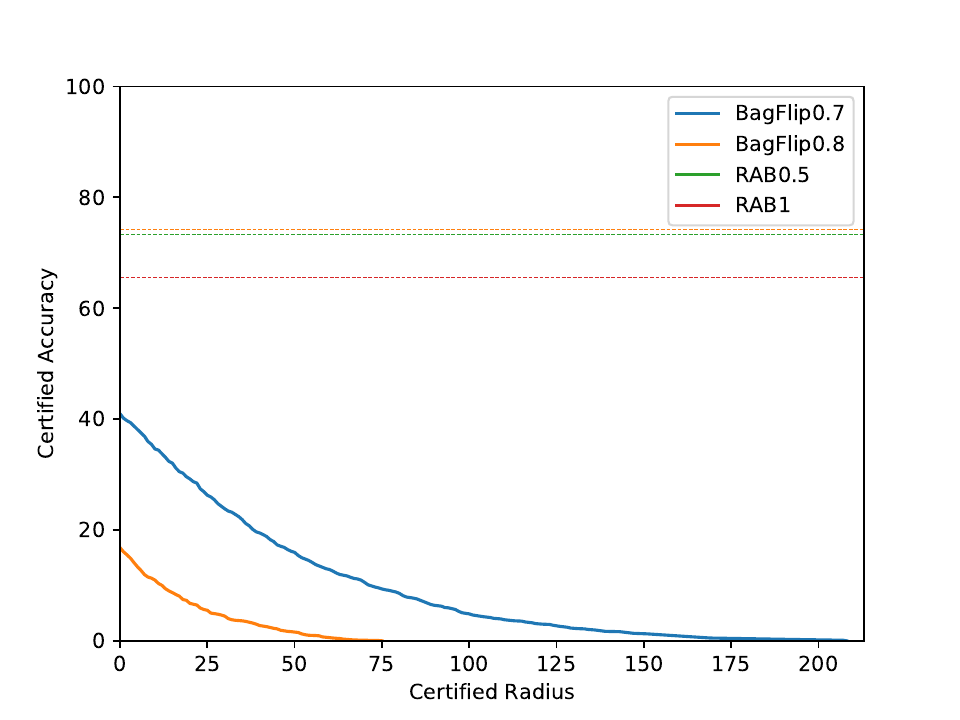}
        \label{fig:rab2}
        }
    \vspace{-1.1em}
    
    \subfigure[Compared to FeatFlip on MNIST-17]{
        \includegraphics[width=.55\textwidth]{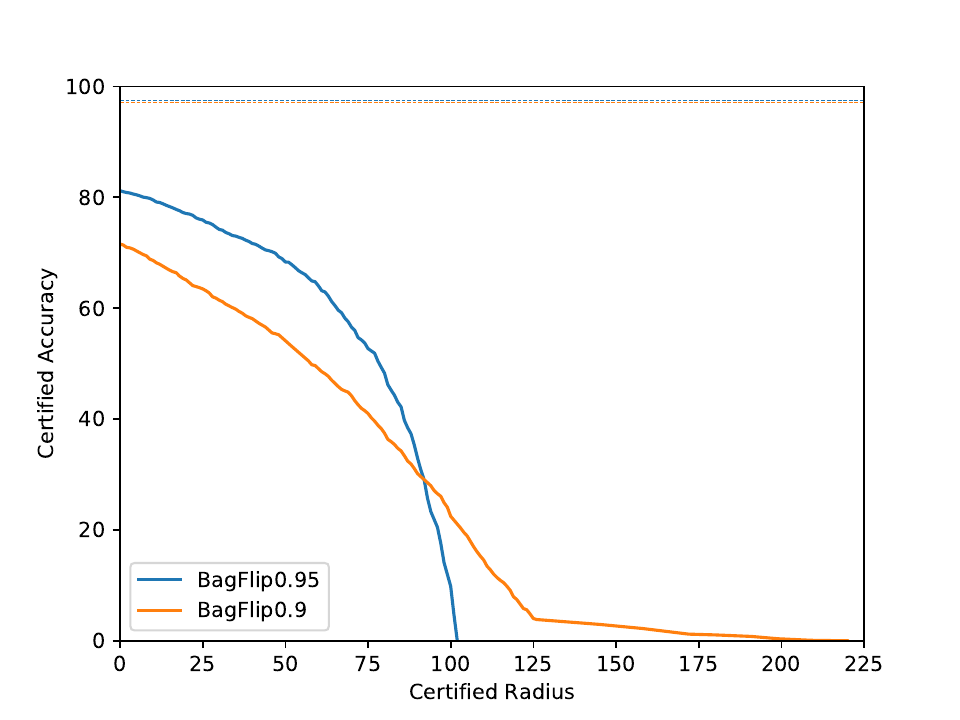}
        }%
\caption{Compared to FeatFlip and RAB.}
\label{fig:feat_rab}
\end{figure}

\paragraph{Comparison to RAB} 
We show the comparison with full configurations of RAB-$\sigma$ in Figures~\ref{fig:rab1}~and~\ref{fig:rab2}, where $\sigma=0.5, 1, 2$ are different Gaussian noise levels.  
Note that RAB's curves are not visible because the certified radius is almost zero anywhere.

\begin{wrapfigure}{R}{0.4\textwidth}
    \centering
\tiny
    
\begin{tikzpicture}
\pgfplotsset{filter discard warning=false}
\pgfplotsset{every axis legend/.append style={
at={(1.5,1)},
anchor=north east}} 

\pgfplotscreateplotcyclelist{whatever}{%
    black,thick,every mark/.append style={fill=blue!80!black},mark=none\\%
    black,thick,dotted,every mark/.append style={fill=red!80!black},mark=none\\%
    }
    \begin{groupplot}[
            group style={
                group size=1 by 1,
                horizontal sep=.05in,
                vertical sep=.05in,
                ylabels at=edge left,
                yticklabels at=edge left,
                xlabels at=edge bottom,
                xticklabels at=edge bottom,
            },
            height=1in,
            xlabel near ticks,
            ylabel near ticks,
            scale only axis,
            width=0.3*\textwidth,
            xmin=0,
            ymin=1,
            ymax=100,
        ]

        \nextgroupplot[
            ylabel=Certifiable Accuracy,
            xmax=0.004,
            cycle list name=whatever,
            xlabel=Poisoning Amount $\assumeradii$ (\%)]
        \addplot table [x=x,y=0, col sep=comma]{data/backdoor/bd_ember.csv};
        \addplot table [x=x,y=0-normal, col sep=comma]{data/backdoor/bd_ember.csv};


    
    \end{groupplot}

\end{tikzpicture}
\caption{\technique-0.95 on EMBER against backdoor attack with $\featflipone$. Dashed lines show normal accuracy.}
  \label{fig:bd_ember}
\end{wrapfigure}

\paragraph{Results on EMBER and CIFAR10}
\technique cannot compute effective certificates for CIFAR10, i.e., the certified accuracy is zero even at $\assumeradii=0$, thus we do not show the figure for CIFAR10.
Figure~\ref{fig:bd_ember} shows the results of \technique on EMBER. 
\technique cannot compute effective certificates for EMBER, neither. 
We leave the improvement of \technique as a future work.

\end{document}